\documentclass[11pt]{article}

\usepackage{acl-style-files/acl}

\usepackage{times}
\usepackage{latexsym}
\usepackage[T1]{fontenc}
\usepackage[utf8]{inputenc}
\usepackage{microtype}
\usepackage{inconsolata}
\usepackage{graphicx}
\usepackage{booktabs}
\usepackage{longtable}
\usepackage{array}
\usepackage[table]{xcolor}
\usepackage{pifont}
\usepackage{amsmath}
\usepackage{multirow}
\usepackage{enumitem}
\usepackage{url}
\usepackage[capitalize,noabbrev]{cleveref}

\newcommand{\ours}{EvoClawBench}
\newcommand{\baseline}{\textsc{Baseline}}
\newcommand{\preskill}{\textsc{PreSkill}}
\newcommand{\postskill}{\textsc{PostSkill}}
\newcolumntype{L}[1]{>{\raggedright\arraybackslash}p{#1}}
\definecolor{promptbg}{RGB}{247,247,247}
\definecolor{promptrule}{RGB}{88,88,88}
\definecolor{yesgreen}{RGB}{30,126,52}
\definecolor{nored}{RGB}{190,50,45}
\definecolor{partialamber}{RGB}{202,137,0}
\newcommand{\yesmark}{\textcolor{yesgreen}{\ding{51}}}
\newcommand{\nomark}{\textcolor{nored}{\ding{55}}}
\newcommand{\partialmark}{\textcolor{partialamber}{\makebox[1em][c]{\ooalign{\hfil\ding{51}\hfil\cr\hfil\ding{55}\hfil\cr}}}}
\newcommand{\promptbox}[3]{%
  \par\noindent
  \begingroup
  \setlength{\fboxsep}{6pt}%
  \colorbox{promptbg}{%
    \begin{minipage}{\dimexpr\linewidth-2\fboxsep\relax}
      {\bfseries #1}\hfill{\footnotesize\ttfamily #2}\par
      \vspace{2pt}{\color{promptrule}\hrule height 0.35pt}\vspace{4pt}
      {\small\ttfamily\raggedright #3\par}
    \end{minipage}%
  }%
  \endgroup
  \par\vspace{0.55\baselineskip}%
}

\title{\ours{}: Can Agents Learn Reusable Skills from Their Own Runs?}

\author{
  Zhiyuan Peng\textsuperscript{1} \quad
  Xin Yin\textsuperscript{2} \quad
  Chenhao Ying\textsuperscript{1} \quad
  Zhe Cui\textsuperscript{3} \\
  \textbf{Zixiang Ding\textsuperscript{3}} \quad
  \textbf{Zhenhua Liu\textsuperscript{3}} \quad
  \textbf{Jiang Wu\textsuperscript{3}} \quad
  \textbf{Yuan Luo\textsuperscript{1}} \\
  \textsuperscript{1}Shanghai Jiao Tong University \quad
  \textsuperscript{2}Zhejiang University \quad
  \textsuperscript{3}Hithink Research \\
  \texttt{\{pzy2000, yingchenhao, yuanluo\}@sjtu.edu.cn} \\
  \texttt{xyin@zju.edu.cn} \\
  \texttt{\{cuizhe, dingzixiang, liuzhenhua, wujiang2\}@myhexin.com}
}

\begin{document}
\maketitle

\begin{abstract}
Existing agent benchmarks primarily test task completion, tool use, or skill utility, but do not isolate whether a runtime can convert evidence from its own runs into reusable skills that improve fresh executions after authoring overhead.
We introduce \textbf{\ours{}}, a benchmark for this closed-loop skill-learning question on repeated, fixture-backed tasks.
\ours{} compares direct execution without skills, \preskill{} authoring before execution, and \postskill{} summarization from first-run evidence followed by a fresh second execution.
The suite contains 100 tasks and 502 sub-problems across coding, data, office, security, operations, and domain-document workflows, with support for multiple agent runtimes.
Experiments with OpenClaw and nanobot under local execution show that direct \baseline{} performance is strongly runtime-dependent: OpenClaw remains below 20\% across models, while nanobot ranges from 56.45\% to 96.13\%.
Self-authored skills have mixed effects.
nanobot \texttt{GPT-5.4} stays above 96\% in all modes and \texttt{MiniMax-M2.7} improves from 90.97\% to 94.50\% under \postskill{}, but nanobot \texttt{DeepSeek-V4-Pro} drops from 77.77\% to 4.80\% with \preskill{} and 0.99\% with \postskill{}.
OpenClaw shows similarly non-monotonic behavior, with some skill runs near baseline and others collapsing.
These results indicate that learning reusable skills from an agent's own runs is selective and cost-sensitive, rather than an automatic benefit of adding skill authoring to an agent loop.
\end{abstract}

\begin{table*}[!t]
\centering
\caption{Comparison of \ours{} with existing benchmarks. ``Skill Cond.'' indicates whether the benchmark includes agent skills. ``Det. Verifier'' indicates whether deterministic (non-LLM) verification is included. ``Closed Loop'' indicates whether the same runtime creates and reuses skills in fresh runs.}
\label{tab:comparison}
\small
\begin{tabular}{lccccc}
\toprule
\textbf{Benchmark} & \textbf{Size} & \textbf{Skill Cond.} & \textbf{Tool Use} & \textbf{Det. Verifier} & \textbf{Closed Loop} \\
\cmidrule(lr){1-6}
SWE-Bench Verified & 500 & None & \yesmark & \yesmark & \nomark \\
Terminal-Bench 2.0 & 89 & None & \yesmark & \yesmark & \nomark \\
WebArena & 812 & None & \yesmark & \partialmark & \nomark \\
AppWorld & 750 & None & \yesmark & \yesmark & \nomark \\
OSWorld & 369 & None & \yesmark & \partialmark & \nomark \\
SkillsBench & 84 & \yesmark & \yesmark & \yesmark & \nomark \\
\midrule
\textbf{\ours{}} & \textbf{100} & \textbf{\yesmark} & \textbf{\yesmark} & \textbf{\yesmark} & \textbf{\yesmark} \\
\bottomrule
\end{tabular}
\end{table*}

\section{Introduction}
\label{sec:intro}

Large language model agents are increasingly used to turn foundation models into interactive workers.
Recent evaluations place agents in repositories, terminals, browsers, graphical interfaces, tool APIs, and executable environments \citep{fourney2024magenticone,wang2025openhands,xu2025agentcompany,agashe2025agents2,cursor2026composer2}.
As a result, the benchmark ecosystem has expanded rapidly.
SWE-Bench and SWE-agent-style evaluations emphasize repository-level software engineering \citep{jimenez2024swebench,yang2024sweagent}; Terminal-Bench, AppWorld, WebArena, and OSWorld measure command-line, app-world, browser, and desktop interaction \citep{zhou2023webarena,trivedi2024appworld,xie2024osworld,merrill2026terminalbench}; and SkillsBench evaluates skills as first-class artifacts across diverse tasks \citep{li2026skillsbench}. These benchmarks have made agent capability more measurable, but they leave a closed-loop learning question under-specified.
They primarily test task completion, tool use, environment interaction, or the utility of curated and self-generated skills.
They do not isolate the process in which the same agent runtime creates a skill, summarizes evidence from its own completed run, reuses that skill in a fresh execution, and pays the full token, cost, and time overhead of doing so.
This leaves a focused evaluation question: \emph{Can LLM agents learn reusable skills from their own runs?}

To answer this question, we introduce \ours{}, a benchmark designed around this question.
\ours{} evaluates an agent on repeated, structured tasks that contain multiple related sub-problems.
This design creates an opportunity for the agent to recognize shared patterns and, in the own-run condition, convert first-run evidence into reusable skills. 
The benchmark focuses on reusable skills because they are becoming a practical way to specialize LLM agents for recurring tool-use tasks.
SkillsBench reports 84{,}192 skills collected within 136 days \citep{li2026skillsbench}, suggesting a shift from prompt engineering by individual users toward a marketplace of reusable procedural memory.
Alongside stronger foundation models and richer harnesses, a complementary line of work treats procedural artifacts as external skills or learned procedures that can be derived, refined, reused, or moved into compact task-family state from agent experience \citep{yang2026autoskill,ni2026trace2skill,wang2026skillx,yang2026skillmaster,xie2026historytostate}.
In this paper, we use \emph{skills} for lightweight, inspectable procedural documents with optional scripts, templates, and references, because such artifacts can be installed, edited, and injected into an agent's context without retraining the model.

For each task, we compare three strategies.
In \baseline{}, the agent solves the task directly and is forbidden from creating or editing skills.
In \preskill{}, the same model and runtime first create one or more task-specific skills, then a fresh execution workspace solves the task using only those generated skills.
In \postskill{}, the agent first solves the task without skills, receives a compact first-run context containing the prompt, grading summary, output previews, and transcript summary, then writes reusable skills that are copied into a fresh second execution. This protocol isolates two related but distinct forms of self-authored skill construction.
\preskill{} is a pre-execution control that measures whether an agent can infer a useful reusable procedure before solving a task.
\postskill{} most directly tests the title question: whether an agent can distill reusable procedural memory from its own completed run.
Both are compared against \baseline{} under execution-only and end-to-end metrics, so a skill workflow must improve not only task score but also justify its additional token, cost, and time overhead.

\begin{figure*}[t]
\centering
\includegraphics[width=\textwidth]{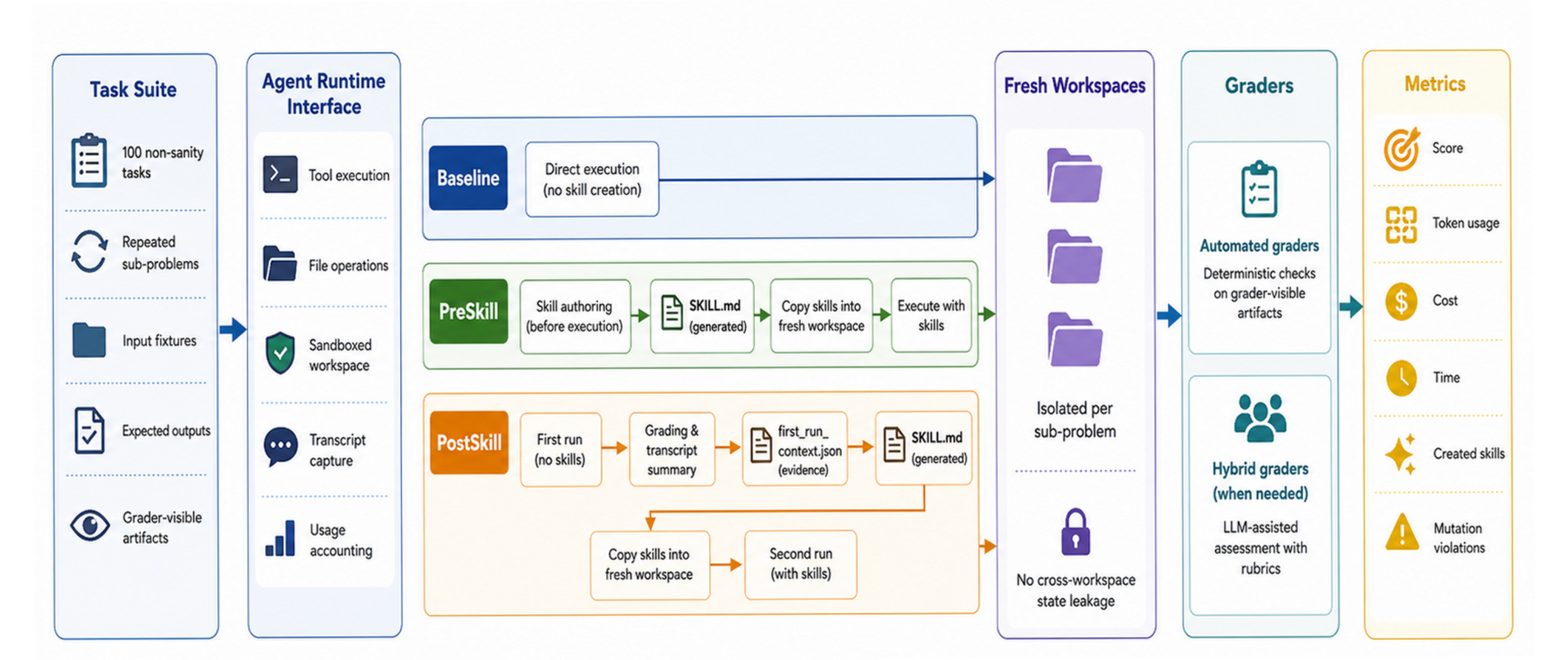}
\caption{Overview for \ours{}. The task-suite panel summarizes the 100 non-sanity tasks in the official suite; \Cref{fig:suite-distribution} breaks down their families and fixture formats. The pipeline routes tasks through a runtime-agnostic agent interface, compares \baseline{}, \preskill{}, and \postskill{} in fresh workspaces, and reports grader-backed quality, resource, skill, and mutation-integrity metrics.}
\label{fig:overview}
\end{figure*}

Our paper makes the following contributions:
\begin{itemize}[leftmargin=*]
    \item \textbf{Benchmark.} We build \ours{}, a benchmark for evaluating whether LLM agents can learn reusable skills from controlled self-authored skill workflows and their own first-run evidence. It covers 100 tasks with repeated sub-problems, fixtures, automated or hybrid graders, and runtime-adapter support.
    \item \textbf{Three-mode evaluation protocol.} We formalize \baseline{}, \preskill{}, and \postskill{} conditions, separating execution-only performance from end-to-end workflow cost and detecting skill mutation violations during reuse.
    \item \textbf{Empirical findings.} We provide a current cross-runtime result table showing that runtime scaffolding strongly affects absolute scores, and that self-authored skills do not produce monotonic gains. Focused subset checks make the cost tradeoff, skill-content controls, and hybrid-judge robustness explicit.
    \item \textbf{Open artifact.} We release the benchmark tasks, fixtures, graders, scripts, and reported result exports at \url{https://github.com/pzy2000/EvoClawBench/tree/anonymous}.
\end{itemize}

\section{Related Work}
\label{sec:related}

\paragraph{Agent and software engineering benchmarks.}
Repository-level benchmarks evaluate whether agents can modify realistic codebases and satisfy executable checks.
SWE-Bench evaluates real GitHub issue resolution \citep{jimenez2024swebench}, while SWE-agent studies agent-computer interfaces for automated software engineering \citep{yang2024sweagent}.
Terminal-Bench focuses on realistic command-line tasks in isolated environments \citep{merrill2026terminalbench}.
MLE-bench extends this line to end-to-end machine-learning engineering on Kaggle competitions \citep{chan2024mlebench}.
AgentBench and GAIA evaluate broader agentic reasoning, tool use, and multi-step problem solving across heterogeneous environments or questions \citep{liu2023agentbench,mialon2023gaia}.
These benchmarks are useful for measuring task completion, but they do not isolate whether an agent can write reusable skills at runtime and then benefit from them.

\paragraph{Web, GUI, and tool-use benchmarks.}
Several benchmarks focus on agents that act through browsers, operating systems, applications, or APIs.
WebArena builds realistic web environments for long-horizon browser tasks \citep{zhou2023webarena}, and Mind2Web provides open-ended web tasks collected from real websites \citep{deng2023mind2web}.
WorkArena++ targets compositional knowledge-work workflows in ServiceNow \citep{boisvert2024workarena}, while OSWorld evaluates multimodal agents on real desktop-computer tasks \citep{xie2024osworld}.
ToolLLM and ToolBench evaluate API-use capability at large scale \citep{qin2023toolllm}.
AppWorld and $\tau$-Bench broaden the scope to controllable app worlds and tool-agent-user interaction settings \citep{trivedi2024appworld,yao2025taubench}.
These environments stress planning and tool execution, but their evaluation conditions do not center on agent-authored artifacts that are reused in fresh runs.

\paragraph{Skill and procedural memory evaluation.}
Skills can be viewed as external procedural memory for agents, related to work on agent skills, reflection, experience accumulation, procedural memory, and open-ended skill acquisition \citep{anthropic2025skills,yao2023react,shinn2023reflexion,zhao2023expel,fang2025memp,wang2023voyager,xu2026agentskills,wu2026agentskillsmemory}.
SkillsBench is closest to our setting because it evaluates skills as first-class artifacts, including curated and self-generated skill conditions \citep{li2026skillsbench}.
Unlike SkillsBench, which evaluates the utility of curated Skills and a pre-solution self-generated Skills condition, our benchmark separates pre-execution skill authoring from post-run skill learning.
In particular, \preskill{} controls for skills authored before task execution, while \postskill{} directly tests whether an agent can convert evidence from its own prior run into a reusable skill for later execution. Taken together, these lines of work motivate task and tool-use evaluation, whereas \ours{} makes the skill lifecycle itself the evaluated object: authoring, own-run distillation, reuse, mutation checks, and end-to-end cost.
\Cref{tab:comparison} summarizes the benchmark position at a high level.
\ours{} is designed to expose whether agent-authored and own-run-derived skills improve task success after accounting for the full cost of creating and reusing skills.

\section{\ours{} Construction}
\label{sec:construction}

\ours{} is constructed to test whether an agent can recognize repeated structure across sub-problems and encode that structure as a reusable skill.
The benchmark implementation is organized around task definitions, workspace fixtures, runtime adapters, graders, and metric aggregation.

\subsection{Task Suite}
\label{subsec:tasks}

\begin{figure*}[t]
\centering
\includegraphics[width=\textwidth]{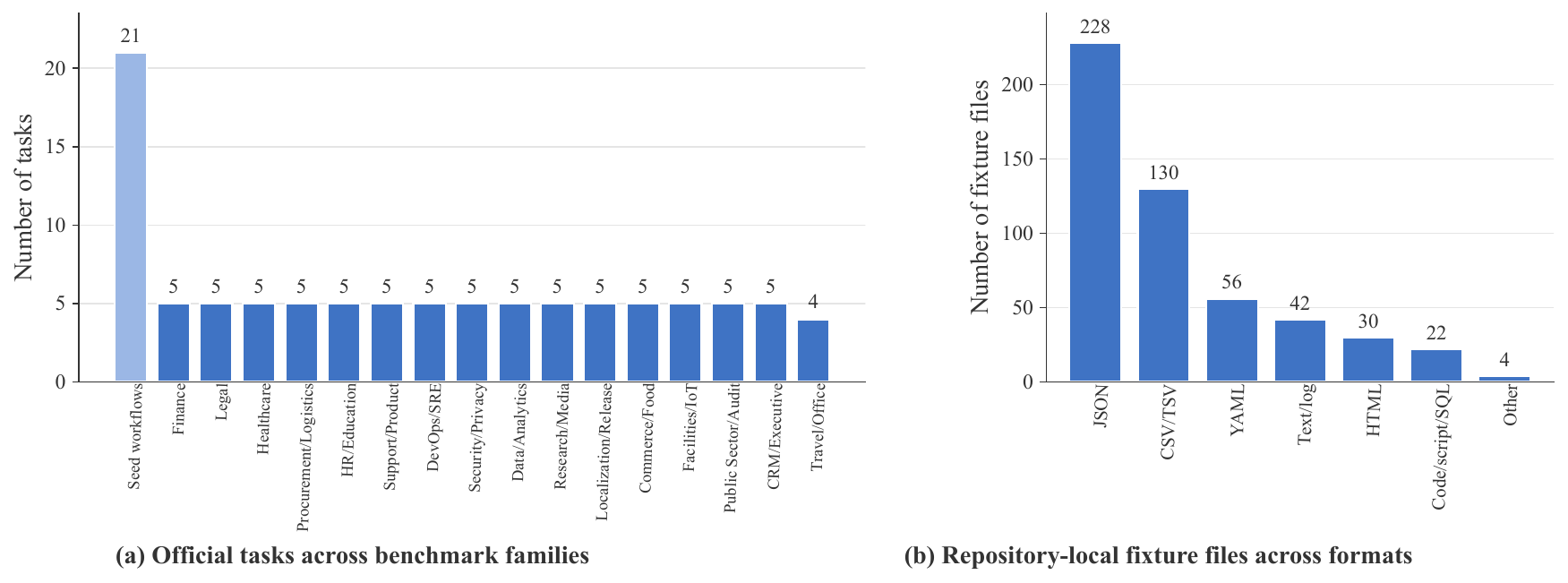}
\caption{Official-suite task and fixture distributions in \ours{}. Panel (a) counts the 100 non-sanity tasks by seed or generated family. Panel (b) counts repository-local workspace fixture files by grouped file format. These metadata counts describe benchmark coverage, not model performance.}
\label{fig:suite-distribution}
\end{figure*}

Tasks are specified as markdown files with YAML frontmatter, a natural-language prompt, expected behavior, sub-problem descriptions, grading criteria, and automated or hybrid grading logic.
Each task typically contains five to ten structurally related sub-problems, creating repeated situations in which an agent can benefit from recording a general procedure rather than solving each instance independently.
The repository also includes a sanity task used by the loader, but we exclude it from official suite counts.
After this exclusion, the suite contains 100 tasks and 502 sub-problems.
It covers recurring workflows such as data transformation, log analysis, API scaffolding, test generation, configuration migration, security review, document extraction, database operations, Excel analytics, web extraction, document generation, data pipelines, email and invoice processing, shell automation, CI generation, dependency audit, environment configuration, and metrics anomaly detection.
The full suite extends these surfaces into harder repository-local families, including finance, legal, healthcare, procurement, HR, support, DevOps/SRE, privacy, localization, commerce, facilities, public-sector audit, CRM, and travel/office coordination.
Many hard-mode fixtures are synthetic repository-local cases with competing evidence packets, stale revisions, and decoy records.
This design reduces privacy risk while preserving repeated, artifact-producing workflows.
\Cref{fig:suite-distribution} summarizes official-suite coverage from task metadata rather than model results: the suite combines seed workflows, generated hard-mode families, and repository-local fixtures in multiple file formats.

\subsection{Workspace and Fixtures}
\label{subsec:workspace}

Each task run is executed in an isolated workspace.
The benchmark copies input assets from its fixture directory into that workspace, and the agent must write grader-visible outputs to the expected paths.
This design avoids relying on conversation-only answers and instead requires concrete artifacts, including JSON files, scripts, reports, Dockerfiles, CI files, and structured extraction outputs.
Task definitions instruct agents not to modify input fixtures, while graders inspect output artifacts rather than trusting the agent's final natural-language answer.

\newpage

The benchmark supports both local subprocess execution and Docker-based execution.
For the reported experiments, we use local subprocess execution through the OpenClaw and nanobot runtimes, set \texttt{--environment local}, run 32 workers, and execute one benchmark run per task and mode.
Because these runs do not use the Docker sandbox, no Docker resource limit is applied.
The reported table should therefore be read as local runtime-adapter results rather than as a claim about all supported sandbox configurations.

\subsection{Skill Authoring and Reuse}
\label{subsec:skills}

\ours{} represents reusable procedures with the standard \texttt{SKILL.md} format adopted by the supported runtimes.
During skill-authoring phases, the workspace is seeded with a \texttt{skill-creator} bundle that instructs the agent to create a well-formed skill.
The benchmark then scans generated skills from the authoring workspace and copies them into fresh execution workspaces for reuse. To prevent a reuse run from becoming another authoring run, \ours{} hashes all non-seeded skill files before and after execution.
If a reuse execution adds, deletes, or edits a skill, the benchmark records a \texttt{skill\_mutation\_violation}.
This guardrail fixes the experimental condition: \preskill{} and \postskill{} execution may consult generated skills, but may not revise them while solving the task.

\subsection{Runtimes}
\label{subsec:runtimes}

\ours{} currently supports OpenClaw and nanobot.
OpenClaw is invoked through \texttt{openclaw agent --message}, while nanobot is invoked through \texttt{nanobot agent --workspace ... --config ... --message ...}.
The runtime abstraction keeps task prompts, workspaces, grading, and metrics shared across claws, allowing the benchmark to compare the same model under different agent scaffolds.
Additional runtimes can be added by implementing the same execution interface and usage extraction logic.

\section{Evaluation Protocol}
\label{sec:protocol}

\begin{figure*}[t]
\centering
\includegraphics[width=\textwidth]{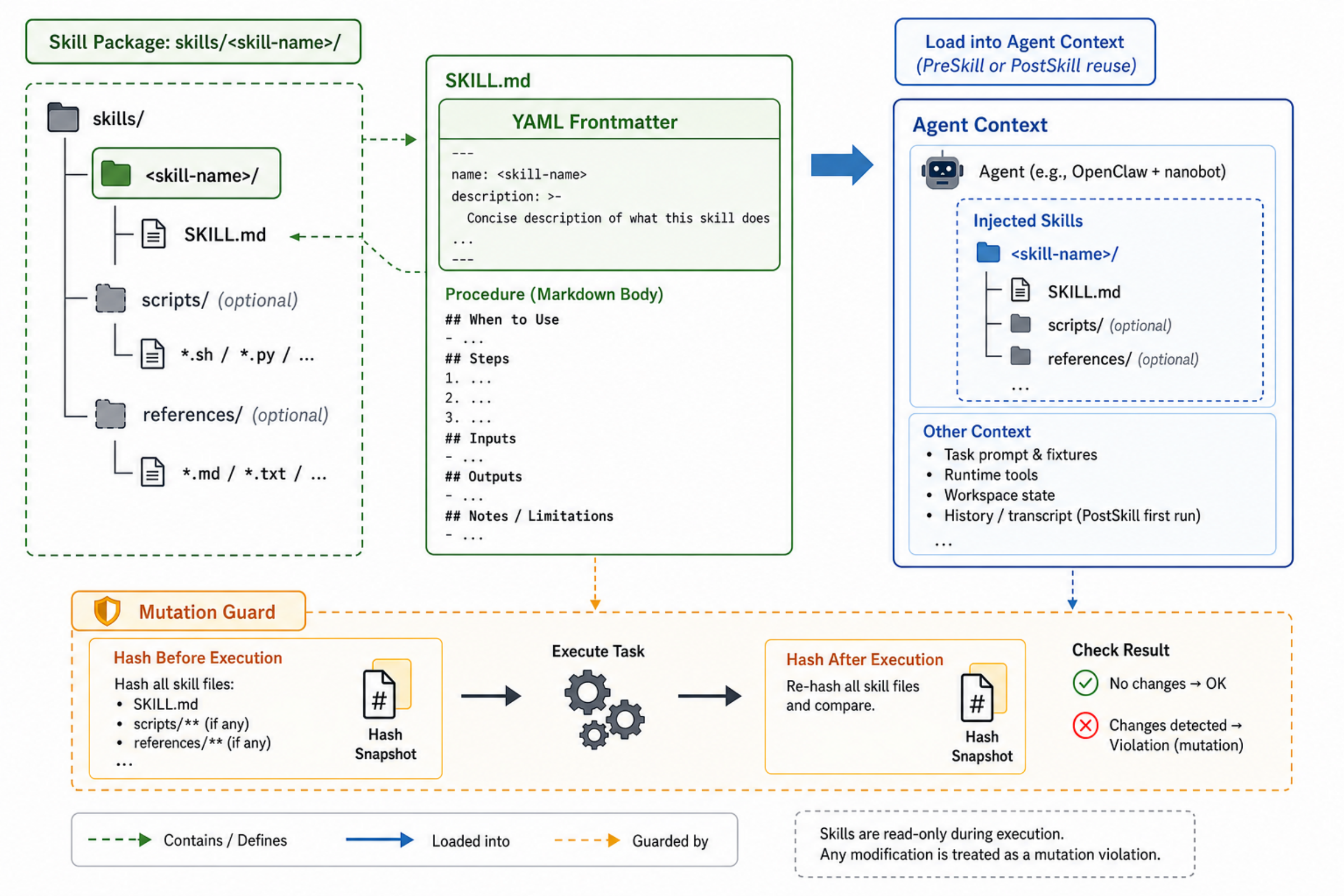}
\caption{Structure of a reusable skill artifact in \ours{}. Generated skills use a \texttt{SKILL.md} document with optional supporting scripts or references, are injected into the agent context during reuse, and are protected by before/after mutation checks during execution.}
\label{fig:skill-anatomy}
\end{figure*}

\subsection{Three Execution Strategies}
\label{subsec:modes}

For a task $t$, model $m$, and runtime $r$, \ours{} evaluates three strategies.

\paragraph{\baseline{}.}
The agent receives the task prompt and solves it directly.
The prompt prefix forbids creating, editing, or deleting skills.
This condition measures the agent's direct task-solving ability.

\paragraph{\preskill{}.}
The agent first enters a skill-authoring phase.
It reads the skill-creator bundle and writes one or more task-specific skills under \texttt{skills/<skill-name>/SKILL.md}.
The final task outputs are not graded during this phase.
The benchmark then creates a fresh workspace, copies the generated skills into it, and runs the task execution phase with skill mutation disabled.

\paragraph{\postskill{}.}
The agent first solves the task in a no-skill execution phase.
The benchmark writes a compact file containing the task prompt, grading details, output summaries, and transcript summaries.
The same runtime and model then summarize reusable skills from that first-run evidence.
A fresh second workspace executes the same task using the summarized skills.

\subsection{Metrics}
\label{subsec:metrics}

Let $S_b(t)$, $S_p(t)$, and $S_q(t)$ denote the mean task scores for \baseline{}, \preskill{}, and \postskill{}, respectively.
The primary execution-only ratios against baseline are:
\begin{equation}
    R_p(t) = \frac{S_p(t)}{S_b(t)}, \qquad
    R_q(t) = \frac{S_q(t)}{S_b(t)}.
\end{equation}
When the baseline score is zero, positive candidate scores are reported as infinite improvement and zero-to-zero comparisons are reported as $1.0$.

\ours{} reports two metric scopes.
\textbf{Execution-only} compares only the task execution phases: direct baseline execution, preskill reuse execution, and postskill second execution.
\textbf{End-to-end} includes all workflow cost: skill generation for \preskill{}, first execution and skill summary for \postskill{}, and direct execution for \baseline{}.

For each scope, the benchmark reports token, cost, and time usage.
Given baseline usage $U_b$ and candidate usage $U_x$, efficiency gain is computed as:
\begin{equation}
    E_x = \frac{U_b}{U_x}.
\end{equation}
Values above $1.0$ therefore indicate that the skill workflow uses fewer resources than the baseline.
The benchmark also reports created-skill counts, heuristic skill quality, and skill mutation violations.
Execution-only metrics test whether a generated skill improves the final task attempt, whereas end-to-end metrics test whether the full workflow justifies its authoring and summarization overhead.
This distinction matters because a skill may preserve task score yet remain unattractive if it substantially increases token use, cost, or wall time.

\section{Results}
\label{sec:results}

\subsection{Experimental Setup}
\label{subsec:setup}

We evaluate five models on the hardened benchmark suite under both OpenClaw and nanobot using local subprocess execution.
All runs use \texttt{uv run scripts/benchmark.py --mode all --workers 32 --environment local}, with \texttt{--runtime openclaw} or \texttt{--runtime nanobot}.
Each task is run once per mode.
Unless otherwise stated, results report the execution-only mean over the benchmark loader's 101 tasks, which include the 100 official tasks plus \texttt{task\_00\_sanity}.
The numerical results in \Cref{tab:main-results} are computed from the repository-local benchmark outputs.

\begin{table*}[t]
\centering
\small
\caption{Repository-local \ours{} results from current OpenClaw and nanobot three-mode outputs. Scores are execution-only mean task scores in percent over the 101-task loader suite; Skills lists \preskill{} / \postskill{} created-skill counts.}
\label{tab:main-results}
\resizebox{\textwidth}{!}{
\begin{tabular}{cccccc}
\toprule
\textbf{Runtime} & \textbf{Model} & \textbf{\baseline{}} & \textbf{\preskill{}} & \textbf{\postskill{}} & \textbf{Skills} \\
\midrule
OpenClaw & \texttt{GPT-5.4} & 18.63 & 18.07 & 1.14 & 22 / 0 \\
OpenClaw & \texttt{Qwen3.6-Plus} & 19.16 & 17.38 & 19.12 & 21 / 22 \\
OpenClaw & \texttt{DeepSeek-V4-Pro} & 19.99 & 15.06 & 0.00 & 21 / 0 \\
OpenClaw & \texttt{MiniMax-M2.7} & 18.88 & 17.00 & 19.10 & 26 / 20 \\
OpenClaw & \texttt{GPT-5.4 mini} & 19.11 & 19.42 & 19.36 & 22 / 20 \\
Nanobot & \texttt{GPT-5.4} & 96.13 & 96.73 & 96.17 & 101 / 100 \\
Nanobot & \texttt{Qwen3.6-Plus} & 56.45 & 59.90 & 54.33 & 70 / 39 \\
Nanobot & \texttt{DeepSeek-V4-Pro} & 77.77 & 4.80 & 0.99 & 27 / 0 \\
Nanobot & \texttt{MiniMax-M2.7} & 90.97 & 92.90 & 94.50 & 99 / 99 \\
Nanobot & \texttt{GPT-5.4 mini} & 88.71 & 87.41 & 89.77 & 100 / 82 \\
\bottomrule
\end{tabular}
}
\end{table*}

\subsection{Evaluation Findings}
\label{subsec:findings}

\paragraph{Finding 1: Runtime scaffolding strongly changes the execution regime.}
Under OpenClaw, direct \baseline{} performance remains below 20\%, ranging from 18.63 for \texttt{GPT-5.4} to 19.99 for \texttt{DeepSeek-V4-Pro}.
Under nanobot, direct \baseline{} scores are substantially higher, ranging from 56.45 for \texttt{Qwen3.6-Plus} to 96.13 for \texttt{GPT-5.4}.
This contrast shows why \ours{} reports runtime as part of the experimental condition rather than treating the model identifier alone as sufficient.
It should not be read as a leaderboard claim that one runtime is inherently superior, because the runtime also determines prompt wrapping, workspace setup, tool invocation, and usage extraction.
The hardened tasks are designed to resist shallow copying of exposed answer fields or schemas: agents must reconcile multi-record evidence, conflicts, revisions, and grader-visible output constraints.
As a result, a generated skill must improve the specific model-runtime execution setting rather than merely preserve performance on an already saturated benchmark.

\paragraph{Finding 2: Skill timing is model-dependent rather than monotonic.}
\texttt{GPT-5.4} illustrates the runtime dependence sharply: OpenClaw falls from 18.63 under \baseline{} to 1.14 under \postskill{}, while nanobot stays above 96\% in all three modes.
\texttt{Qwen3.6-Plus} improves under nanobot \preskill{} but regresses under nanobot \postskill{}, whereas its OpenClaw row remains near 19\% for \baseline{} and \postskill{}.
\texttt{GPT-5.4 mini} improves under OpenClaw skill workflows and under nanobot \postskill{}, but its nanobot \preskill{} score is slightly below its nanobot \baseline{}.
These rows do not support a blanket claim that \preskill{} consistently outperforms \postskill{}, or that either skill workflow reliably improves over direct execution across runtimes.

\paragraph{Finding 3: Skill workflows still add substantial overhead.}
Even when execution-only quality is similar, skill workflows include authoring or summarization phases.
For \texttt{Qwen3.6-Plus}, end-to-end token-efficiency ratios are 0.38 for \preskill{} and 0.30 for \postskill{}, meaning the full workflows use substantially more tokens than direct \baseline{} execution.
For \texttt{MiniMax-M2.7}, the corresponding ratios are 0.21 and 0.26; for \texttt{GPT-5.4 mini}, they are 0.32 and 0.28.
For \texttt{GPT-5.4}, \preskill{} has an end-to-end token-efficiency ratio of 0.36; for \texttt{DeepSeek-V4-Pro}, the corresponding ratio is 0.40.
The current evidence therefore suggests that runtime skill creation must be selective: skill creation needs to clear both a quality bar and an amortized-cost bar.

\begin{figure*}[t]
\centering
\includegraphics[width=\textwidth]{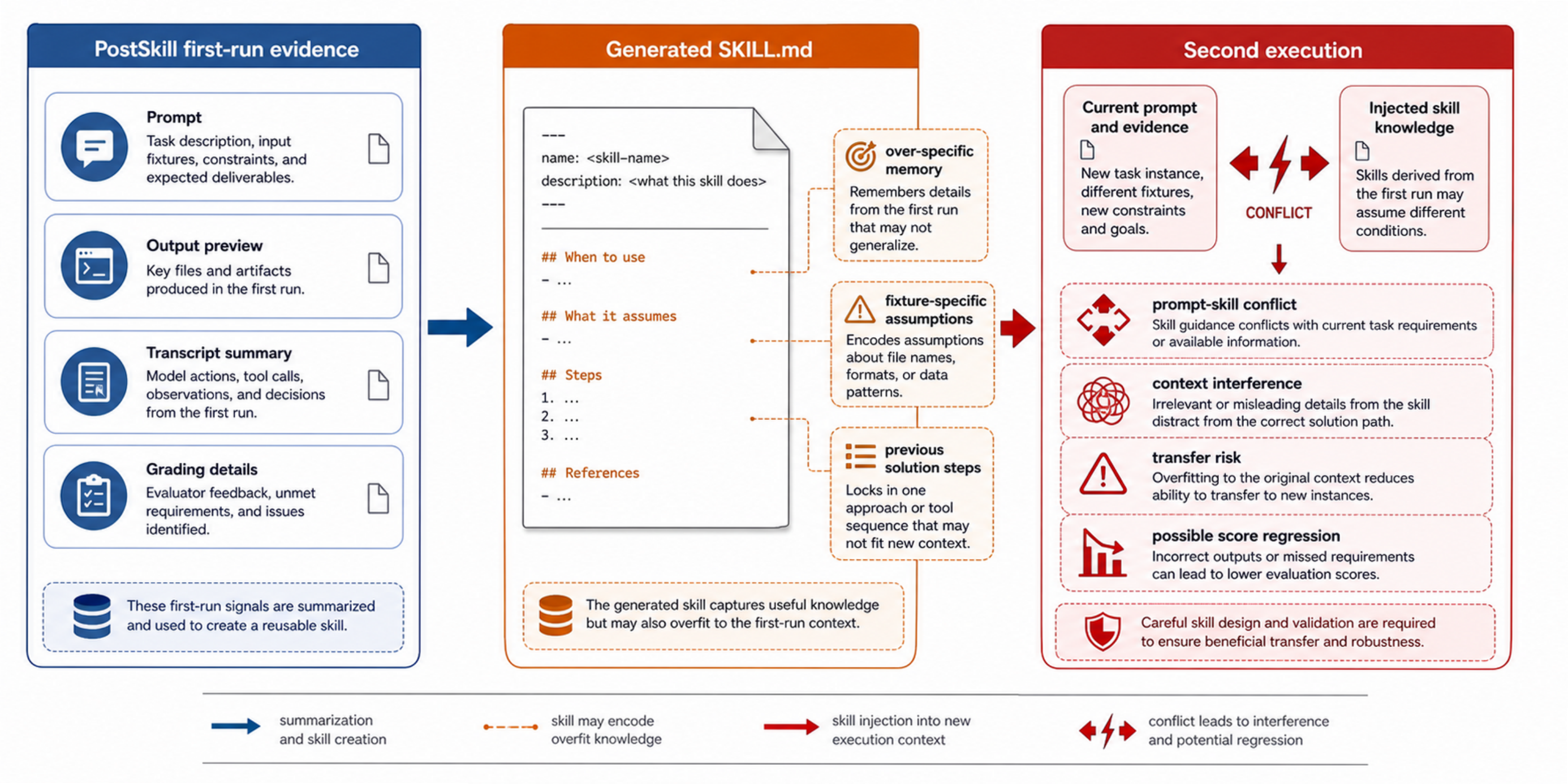}
\caption{Qualitative failure mode for experience-derived skills. In \postskill{}, first-run evidence can help produce reusable procedural memory, but it can also encode over-specific assumptions that interfere with the second execution context.}
\label{fig:failure-case}
\end{figure*}

\paragraph{Finding 4: Skill counts are not sufficient to predict gains.}
The OpenClaw rows create between 21 and 26 \preskill{} skills, but their \preskill{} scores range from 15.06 to 19.42.
The nanobot rows create many more \preskill{} skills, yet the score range remains wide.
\postskill{} skill counts are also not enough to explain outcomes: rows with many \postskill{} skills can land below, near, or above their own \baseline{} scores.
This suggests that future evaluations should inspect generated skill content and reuse behavior, not only the number of produced skill directories.
Most observed regressions are therefore better explained by the quality and fit of generated skills than by the raw number of skill directories.
\Cref{fig:failure-case} illustrates why this inspection matters for \postskill{}.
The summary phase receives compact first-run evidence rather than hidden state, but that evidence can still make a generated skill too specific to one execution context.
When reused in a fresh workspace, such a skill may preserve an incidental assumption, route attention to the wrong fixture pattern, or skip checks that the second execution still requires.
The qualitative case therefore complements the aggregate skill-count result: the question is not only whether a skill exists, but whether the procedure it encodes remains valid under reuse.

\subsection{Subset Cost, Amortization, and Robustness Checks}
\label{subsec:subset-checks}

To address cost and judge-sensitivity concerns directly, we ran focused subset experiments on the 12 hybrid tasks using OpenClaw and \texttt{GPT-5.4 mini}.
\Cref{tab:subset-cost} shows the end-to-end resource tradeoff.
On this subset, both skill workflows slightly reduce score while increasing tokens, cost, and wall-clock time.
The negative score-per-extra-token values therefore make the cost-quality tradeoff explicit rather than hiding it behind execution-only scores.
The subset is not a replacement for the full-suite table; it is a targeted audit of the hybrid tasks where judge cost and scoring robustness are most salient.

\Cref{tab:subset-break-even} estimates how many repeated reuses would be needed before the authoring or summary overhead is amortized.
For tokens and provider cost, neither \preskill{} nor \postskill{} breaks even on this subset because the reuse executions are not cheaper than direct execution.
Wall-clock time can break even after repeated reuse, but only after 14 \preskill{} executions or 9 \postskill{} executions under the observed local run times.
Thus the subset supports a cost-sensitive interpretation: skills may be worth keeping only when they are reused many times and preserve or improve quality.

We also ran a 4-task ablation subset (\texttt{task\_02}, \texttt{task\_07}, \texttt{task\_15}, and \texttt{task\_21}) to separate skill content from reuse scaffolding.
\Cref{tab:subset-ablation} shows that normal \preskill{} reuse improves the mean score by 4.05 points, but an empty-skill scaffold improves by 19.84 points, while an irrelevant skill improves by only 1.52 points.
This pattern is not evidence that empty skills are generally useful; rather, it shows that the reuse prompt, context reset, and run-to-run variance can explain some gains that might otherwise be attributed to skill content.
The \postskill{} comparison is also cautious: normal \postskill{} reuse is slightly above baseline, but discarding the generated skill after the summary phase falls below baseline.
Because these ablations are single subset runs rather than repeated-seed estimates, we use them as diagnostic controls for attribution, not as a new headline result.

Finally, we regraded the 36 task-mode outputs from the 12 hybrid-task subset with alternate judges.
Because hybrid scores combine deterministic checks with an LLM judgment, the regrade holds the automated sub-scores fixed and replaces only the LLM-judge component.
One alternate judge shows high agreement with the original scores: Pearson correlation is 0.9970, mean absolute score delta is 0.0141, 35 of 36 pairs are within 0.05, and all 36 pairs are within 0.10.
The largest disagreement is \texttt{task\_18\_dep\_audit} in \preskill{} (+0.0880).
As a second check, \texttt{GPT-5.4 mini} also preserves the overall trend, though less tightly: Pearson correlation is 0.9512, mean absolute score delta is 0.0609, and 30 of 36 pairs are within 0.10.
The largest \texttt{GPT-5.4 mini} disagreement is \texttt{task\_11\_web\_extraction} in \baseline{} (-0.3972).
These subset audits reduce the concern that the reported hybrid-task pattern is an artifact of a single judge, while the remaining disagreements still motivate auditing high-stakes leaderboard claims.
They do not remove judge dependence as a limitation; instead, they show how future benchmark releases should report both original and alternate-judge agreement when hybrid grading affects a result.

\section{Conclusion}
\label{sec:discussion}

\ours{} evaluates the full lifecycle of procedural memory: skill creation, own-run skill distillation, reuse, and library maintenance.
The current results suggest that agents are not yet reliable at this lifecycle: neither the pre-execution \preskill{} control nor the own-run \postskill{} condition is uniformly dominant, and \Cref{fig:failure-case} shows how own-run summaries can overfit a second execution context.
Future skill systems should use selective creation policies and validation before reuse, while benchmark reports should include provenance, mutation checks, and end-to-end resources.

\section*{Limitations}

\paragraph{Transfer scope.}
The v1 protocol evaluates \postskill{} by rerunning the same task with the same fixtures, which measures within-task skill distillation but does not yet test transfer to unseen related tasks.
This means the reported results should not be interpreted as evidence that generated skills generalize across new tasks, users, or domains.

\paragraph{Experimental coverage.}
The task suite covers many practical office, data, coding, DevOps, security, and domain-document workflows, but it is still smaller than large public agent benchmarks.
The current reported table covers two runtimes under one local execution configuration with one run per task and mode.
Repeated seeds, additional models, confidence intervals, Docker resource limits, and more runtime replications remain future work.

\paragraph{Grading dependence.}
Hybrid grading tasks may depend on the selected judge model, although automated checks are used whenever feasible.
Future result files should store the judge model explicitly because absolute scores may change under a different judge even if the task prompts and artifacts are unchanged.

\paragraph{Runtime scope.}
The current implementation and reported experiments cover OpenClaw and nanobot.
Broader runtime coverage is still needed before claiming conclusions across a wider set of agent runtimes.

\section*{Ethical Considerations}

\ours{} is intended as a research benchmark for evaluating agent skill authoring, own-run skill distillation, and reuse, not as a deployment recipe for autonomous skill libraries.
The benchmark uses repository-local task fixtures and synthetic examples whenever sensitive domains are represented; it does not involve human subjects, crowdworkers, or newly collected personal data.
Several tasks simulate workflows involving privacy, security, legal, healthcare, finance, or public-sector records, but these are benchmark scenarios for controlled evaluation rather than operational recommendations. The main risk is dual use: better skill creation from an agent's own runs could make agents more effective at legitimate repeated work, but could also preserve unsafe procedures, overfit to private context, or automate harmful operational tasks if deployed without review.
The benchmark therefore records skill mutation violations and separates execution-only quality from end-to-end cost, but it does not solve policy questions about which generated skills should be installed, shared, or trusted.
Systems using self-authored skills should include human review, privacy checks, provenance tracking, and mechanisms for disabling or deleting unsafe skills.

\section*{Artifacts and Data Statement}

The benchmark artifact used for this manuscript is the repository-local \texttt{evoclawbench/} package.
It contains task definitions under \texttt{tasks/}, input fixtures under \texttt{assets/}, grading and metric code under \texttt{scripts/}, and reported OpenClaw and nanobot result exports under \texttt{results/}.
The project metadata in \texttt{pyproject.toml} declares the package license as MIT.
For anonymous review, the artifact is available at \url{https://anonymous.4open.science/r/EvoClawBench-9380/}.
The reported task fixtures are benchmark cases rather than newly collected human-subject data, and the manuscript does not rely on private external datasets.

\bibliography{main}

\clearpage
\onecolumn
\appendix

\section{Mode Prompt Prefixes}
\label{app:prefixes}
\ours{} controls each execution condition by prepending a mode-specific prefix to the task prompt loaded from the task markdown file. The prefix text below reproduces the benchmark runtime helper wording; trailing blank lines are omitted and long lines are wrapped by LaTeX for typesetting. The assembled prompt is always the selected prefix followed by the task prompt; task prompts themselves remain stored in \path{evoclawbench/tasks/task_*.md}.

\promptbox{Baseline execution prefix}{\detokenize{BASELINE MODE:}}{%
\detokenize{Complete the task directly. You must NOT create, edit, or delete any skills or SKILL.md files. Solve each sub-problem independently from scratch, and focus on producing the required grader-visible outputs.}}

\promptbox{Preskill authoring prefix}{\detokenize{PRESKILL AUTHOR MODE:}}{%
\detokenize{Before solving the task, create a task-specific reusable skill. Read `skills/skill-creator/SKILL.md` and write one or more new skills under `skills/<skill-name>/SKILL.md` that would help an agent solve this task later. Do not produce the final task outputs in `outputs/`; this phase is only for skill authoring. Do not replace or delete `skills/skill-creator`.}}

\promptbox{Postskill summary prefix}{\detokenize{POSTSKILL SUMMARY MODE:}}{%
\detokenize{Summarize a reusable task-specific skill from a completed first run. Read `.evoclawbench/first_run_context.json` for the first run prompt, grading details, workspace output summary, and transcript summary. Create one or more skills under `skills/<skill-name>/SKILL.md` so a later agent can solve the same task more accurately and efficiently. Do not redo the task and do not write final task outputs in `outputs/`. Do not replace or delete `skills/skill-creator`.}}

\promptbox{Skill reuse execution prefix}{\detokenize{SKILL REUSE EXECUTION MODE:}}{%
\detokenize{Complete the task using the existing skills in `skills/` when helpful. You must NOT create, edit, or delete any skills or SKILL.md files during this execution phase. Focus on producing the required grader-visible outputs and validate them before finishing.}}

\begin{center}
\small
\setlength{\tabcolsep}{3pt}
\captionof{table}{Prompt assembly by benchmark phase.}
\label{tab:appendix-mode-prompts}
\begin{tabular}{L{0.17\textwidth}L{0.19\textwidth}L{0.26\textwidth}L{0.28\textwidth}}
\toprule
\textbf{Mode} & \textbf{Phase} & \textbf{Prompt assembly} & \textbf{Graded behavior} \\
\midrule
baseline & baseline execution & Baseline prefix + task prompt & No skill creation or mutation; grade final outputs. \\
preskill\_author & preskill authoring & Preskill author prefix + task prompt & Seed skill-creator; write skills; final task outputs are not graded. \\
preskill\_execute & preskill reuse execution & Skill reuse prefix + task prompt & Copy generated skills into a fresh workspace; forbid skill mutation. \\
postskill\_first & postskill first execution & Baseline prefix + task prompt & Solve without skills; grade outputs and write first-run context. \\
postskill\_summary & postskill summary & Postskill summary prefix + task prompt & Read first-run context; write reusable skills; do not redo final outputs. \\
postskill\_second & postskill reuse execution & Skill reuse prefix + task prompt & Copy summarized skills into a fresh workspace; forbid skill mutation. \\
\bottomrule
\end{tabular}
\end{center}

\section{PostSkill First-Run Context}
\label{app:first-run-context}
In \postskill{}, the first execution is graded before any summary skill is written. The benchmark then writes \path{.evoclawbench/first_run_context.json} into the first workspace and copies that file into the summary workspace. This prevents the summary phase from relying on hidden state while still giving it compact evidence from the completed attempt.
\begin{center}
\small
\captionof{table}{First-run context fields serialized for the summary phase.}
\label{tab:first-run-context-fields}
\begin{tabular}{L{0.24\textwidth}L{0.66\textwidth}}
\toprule
\textbf{Field} & \textbf{Purpose} \\
\midrule
task\_id & Stable task identifier copied from the task markdown frontmatter. \\
task\_name & Human-readable task name. \\
prompt & The original task prompt, without the mode prefix. \\
expected\_behavior & The task-level expected behavior section. \\
grading\_criteria & Parsed checklist items from the grading criteria section. \\
grade & Serialized grade result from the first no-skill execution. \\
outputs & Small summaries of files produced under the first-run workspace outputs directory. \\
transcript\_summary & A compact role/text transcript summary capped by the benchmark helper. \\
\bottomrule
\end{tabular}
\end{center}
The summary phase receives the same task prompt after the \postskill{} summary prefix, but it is instructed not to redo the task and not to write final task outputs under \path{outputs/}. Skills produced in the summary workspace are copied into a fresh second execution workspace, where mutation checks are applied before and after execution.

\section{Subset Audit Details}
\label{app:subset-audits}
The following tables provide the detailed subset evidence summarized in \Cref{subsec:subset-checks}.

\begin{center}
\small
\captionof{table}{End-to-end cost on the 12-task hybrid subset. Scores are percentages. ``Delta/1M tok.'' is score-point change per one million extra tokens relative to \baseline{}.}
\label{tab:subset-cost}
\resizebox{\textwidth}{!}{
\begin{tabular}{lrrrrrrr}
\toprule
\textbf{Mode} & \textbf{Score} & \textbf{Delta} & \textbf{Tokens} & \textbf{USD} & \textbf{Time h} & \textbf{Requests} & \textbf{Delta/1M tok.} \\
\midrule
\baseline{} & 80.96 & 0.00 & 1.94M & 1.262 & 0.305 & 84 & -- \\
\preskill{} & 79.77 & -1.19 & 7.07M & 3.745 & 0.689 & 270 & -0.232 \\
\postskill{} & 79.21 & -1.75 & 7.71M & 4.128 & 0.833 & 306 & -0.303 \\
\bottomrule
\end{tabular}
}
\end{center}

\begin{center}
\small
\captionof{table}{Amortized break-even reuse count on the same subset. ``Never'' means the reuse execution is not cheaper than direct \baseline{} execution for that resource.}
\label{tab:subset-break-even}
\begin{tabular}{lccc}
\toprule
\textbf{Mode} & \textbf{Tokens} & \textbf{USD} & \textbf{Time} \\
\midrule
\preskill{} & never & never & 14 \\
\postskill{} & never & never & 9 \\
\bottomrule
\end{tabular}
\end{center}

\begin{center}
\small
\captionof{table}{Skill-content ablation on a 4-task representative subset. Scores and deltas are percentages; resources are end-to-end for each variant.}
\label{tab:subset-ablation}
\resizebox{\textwidth}{!}{
\begin{tabular}{lrrrrr}
\toprule
\textbf{Variant} & \textbf{Score} & \textbf{Delta} & \textbf{Tokens} & \textbf{USD} & \textbf{Requests} \\
\midrule
\baseline{} & 54.24 & 0.00 & 0.80M & 0.236 & 35 \\
\preskill{} normal & 58.28 & 4.05 & 2.80M & 0.658 & 102 \\
Empty scaffold & 74.08 & 19.84 & 0.54M & 0.196 & 24 \\
Irrelevant skill & 55.76 & 1.52 & 0.92M & 0.241 & 38 \\
\postskill{} normal & 55.70 & 1.47 & 2.79M & 0.810 & 108 \\
Summary, no skill & 50.01 & -4.23 & 2.59M & 0.743 & 102 \\
\bottomrule
\end{tabular}
}
\end{center}

\section{Official Task Inventory}
\label{app:task-inventory}
The loader currently finds 101 task files. The official benchmark size reported in the paper excludes \texttt{task\_00\_sanity}, leaving 100 tasks and 502 sub-problems. The table uses ASCII-safe names because the review build uses \texttt{pdflatex} without CJK font configuration.
\begingroup
\footnotesize
\setlength{\tabcolsep}{1pt}
\begin{longtable}{L{0.09\textwidth}L{0.30\textwidth}L{0.25\textwidth}L{0.10\textwidth}L{0.07\textwidth}L{0.07\textwidth}L{0.08\textwidth}}
\caption{Official task inventory, excluding \texttt{task\_00\_sanity}.}\label{tab:task-inventory}\\
\toprule
\textbf{Task} & \textbf{Name} & \textbf{Family/category} & \textbf{Grade} & \textbf{Sub.} & \textbf{Files} & \textbf{Ext.} \\
\midrule
\endfirsthead
\toprule
\textbf{Task} & \textbf{Name} & \textbf{Family/category} & \textbf{Grade} & \textbf{Sub.} & \textbf{Files} & \textbf{Ext.} \\
\midrule
\endhead
task\_01 & Batch Data Transform & Seed / data transformation & automated & 6 & 6 & csv, json, tsv, xml, yaml \\
task\_02 & Log Analysis & Seed / data analysis & hybrid & 5 & 5 & log \\
task\_03 & API Integration Scaffold & Seed / code generation & automated & 5 & 5 & json \\
task\_04 & Test Generation & Seed / testing & automated & 6 & 6 & py \\
task\_05 & Config Migration v1 to v2 & Seed / data transformation & automated & 5 & 5 & json \\
task\_06 & Security Code Review & Seed / code analysis & hybrid & 5 & 5 & go, js, py \\
task\_07 & Document Data Extraction & Seed / data extraction & hybrid & 5 & 5 & txt \\
task\_08 & Database Schema Operations & Seed / database & automated & 5 & 5 & sql \\
task\_09 & Excel Analytics Report Generation & Seed / data transformation & automated & 5 & 5 & csv \\
task\_10 & Git Repository History Analysis & Seed / data analysis & automated & 5 & 5 & sh \\
task\_11 & Web Page Structured Data Extraction & Seed / data extraction & hybrid & 5 & 5 & html \\
task\_12 & Word Document Generation & Seed / code generation & automated & 5 & 5 & json \\
task\_13 & Multi-Dataset Statistical Analysis Pipeline & Seed / data analysis & hybrid & 5 & 15 & csv \\
task\_14 & Enterprise Email Thread Analysis & Seed / office automation & hybrid & 5 & 5 & json \\
task\_15 & Shell Automation Script Generation & Seed / automation & hybrid & 5 & 5 & yaml \\
task\_16 & Dockerfile and CI Pipeline Generation & Seed / devops & hybrid & 5 & 5 & json \\
task\_17 & Enterprise Invoice Expense Processing & Seed / office daily & hybrid & 5 & 5 & csv, json, tsv, txt, xml \\
task\_18 & Dependency Security Audit & Seed / security & hybrid & 5 & 5 & json, mod, none, txt, xml \\
task\_19 & Meeting Notes Structured Extraction & Seed / office daily & hybrid & 5 & 5 & txt \\
task\_20 & Multi-Environment Config Generation & Seed / devops automation & automated & 5 & 5 & json \\
task\_21 & System Metrics Anomaly Detection & Seed / data analysis & hybrid & 5 & 5 & csv \\
task\_22 & Finance Ledger Reconciliation & Finance & automated & 5 & 5 & csv \\
task\_23 & Subscription Revenue Audit & Finance & automated & 5 & 5 & csv \\
task\_24 & Bank KYC Packet Review & Finance & automated & 5 & 5 & csv \\
task\_25 & Loan Document Checklist & Finance & automated & 5 & 5 & csv \\
task\_26 & Tax Form Consistency & Finance & automated & 5 & 5 & csv \\
task\_27 & Contract Clause Extraction & Legal & automated & 5 & 5 & txt \\
task\_28 & Discovery Document Tagging & Legal & automated & 5 & 5 & txt \\
task\_29 & Policy Exception Mapping & Legal & automated & 5 & 5 & txt \\
task\_30 & Nda Obligation Review & Legal & automated & 5 & 5 & txt \\
task\_31 & Vendor Contract Risk & Legal & automated & 5 & 5 & txt \\
task\_32 & Appointment Referral Triage & Healthcare & automated & 5 & 5 & json \\
task\_33 & Pharmacy Inventory Reorder & Healthcare & automated & 5 & 5 & json \\
task\_34 & Clinical Trial Eligibility & Healthcare & automated & 5 & 5 & json \\
task\_35 & Insurance Prior Auth Review & Healthcare & automated & 5 & 5 & json \\
task\_36 & Lab Result Followup Queue & Healthcare & automated & 5 & 5 & json \\
task\_37 & Procurement Bid Scoring & Procurement/Logistics & automated & 5 & 5 & yaml \\
task\_38 & Purchase Order Three Way Match & Procurement/Logistics & automated & 5 & 5 & yaml \\
task\_39 & Shipping Exception Resolution & Procurement/Logistics & automated & 5 & 5 & yaml \\
task\_40 & Warehouse Picklist Validation & Procurement/Logistics & automated & 5 & 5 & yaml \\
task\_41 & Supplier Risk Packet & Procurement/Logistics & automated & 5 & 5 & yaml \\
task\_42 & Employee Onboarding Checklist & HR/Education & automated & 5 & 5 & json \\
task\_43 & Interview Feedback Calibration & HR/Education & automated & 5 & 5 & json \\
task\_44 & Admissions Packet Screening & HR/Education & automated & 5 & 5 & json \\
task\_45 & Assignment Rubric Grading & HR/Education & automated & 5 & 5 & json \\
task\_46 & Training Completion Audit & HR/Education & automated & 5 & 5 & json \\
task\_47 & Support Ticket Escalation & Support/Product & automated & 5 & 5 & json \\
task\_48 & Kb Gap Analysis & Support/Product & automated & 5 & 5 & json \\
task\_49 & App Review Theme Mining & Support/Product & automated & 5 & 5 & json \\
task\_50 & Bug Report Deduplication & Support/Product & automated & 5 & 5 & json \\
task\_51 & Call Center Quality Review & Support/Product & automated & 5 & 5 & json \\
task\_52 & Kubernetes Policy Review & DevOps/SRE & automated & 5 & 5 & yaml \\
task\_53 & Terraform Plan Drift & DevOps/SRE & automated & 5 & 5 & yaml \\
task\_54 & SLO Burn Rate Analysis & DevOps/SRE & automated & 5 & 5 & yaml \\
task\_55 & CI Pipeline Hardening & DevOps/SRE & automated & 5 & 5 & yaml \\
task\_56 & Backup Restore Drill & DevOps/SRE & automated & 5 & 5 & yaml \\
task\_57 & Security Alert Correlation & Security/Privacy & automated & 5 & 5 & json \\
task\_58 & Phishing Report Triage & Security/Privacy & automated & 5 & 5 & json \\
task\_59 & Vulnerability Exception Review & Security/Privacy & automated & 5 & 5 & json \\
task\_60 & PII Redaction Release & Security/Privacy & automated & 5 & 5 & json \\
task\_61 & DSR Request Routing & Security/Privacy & automated & 5 & 5 & json \\
task\_62 & SQL Schema Migration Review & Data/Analytics & automated & 5 & 5 & csv \\
task\_63 & Messy Csv Normalization & Data/Analytics & automated & 5 & 5 & csv \\
task\_64 & Dashboard Metric Reconciliation & Data/Analytics & automated & 5 & 5 & csv \\
task\_65 & Data Quality Rule Authoring & Data/Analytics & automated & 5 & 5 & csv \\
task\_66 & Experiment Results Analysis & Data/Analytics & automated & 5 & 5 & csv \\
task\_67 & Research Claim Evidence & Research/Media & automated & 5 & 5 & html \\
task\_68 & Newsroom Fact Check & Research/Media & automated & 5 & 5 & html \\
task\_69 & Literature Table Extraction & Research/Media & automated & 5 & 5 & html \\
task\_70 & Public Web Directory Extraction & Research/Media & automated & 5 & 5 & html \\
task\_71 & Citation Deduplication & Research/Media & automated & 5 & 5 & html \\
task\_72 & Localization Placeholder Qa & Localization/Release & automated & 5 & 5 & json \\
task\_73 & Glossary Compliance Review & Localization/Release & automated & 5 & 5 & json \\
task\_74 & Release Notes Curation & Localization/Release & automated & 5 & 5 & json \\
task\_75 & Changelog Impact Matrix & Localization/Release & automated & 5 & 5 & json \\
task\_76 & Feature Flag Cleanup Plan & Localization/Release & automated & 5 & 5 & json \\
task\_77 & Ecommerce Catalog Normalization & Commerce/Food & automated & 5 & 5 & csv \\
task\_78 & Retail Returns Root Cause & Commerce/Food & automated & 5 & 5 & csv \\
task\_79 & Restaurant Inspection Summary & Commerce/Food & automated & 5 & 5 & csv \\
task\_80 & Food Delivery Refund Review & Commerce/Food & automated & 5 & 5 & csv \\
task\_81 & Marketplace Listing Policy & Commerce/Food & automated & 5 & 5 & csv \\
task\_82 & Facilities Maintenance Prioritization & Facilities/IoT & automated & 5 & 5 & json \\
task\_83 & Fleet Maintenance Schedule & Facilities/IoT & automated & 5 & 5 & json \\
task\_84 & Iot Sensor Anomaly Review & Facilities/IoT & automated & 5 & 5 & json \\
task\_85 & Smart Home Support Diagnosis & Facilities/IoT & automated & 5 & 5 & json \\
task\_86 & Robotics Workcell Event Report & Facilities/IoT & automated & 5 & 5 & json \\
task\_87 & Civic Service Request Routing & Public Sector/Audit & automated & 5 & 5 & json \\
task\_88 & Public Records Redaction & Public Sector/Audit & automated & 5 & 5 & json \\
task\_89 & Grant Application Completeness & Public Sector/Audit & automated & 5 & 5 & json \\
task\_90 & Audit Evidence Collection & Public Sector/Audit & automated & 5 & 5 & json \\
task\_91 & Risk Register Rollup & Public Sector/Audit & automated & 5 & 5 & json \\
task\_92 & CRM Pipeline Hygiene & CRM/Executive & automated & 5 & 5 & csv \\
task\_93 & Sales Forecast Variance & CRM/Executive & automated & 5 & 5 & csv \\
task\_94 & Marketing Campaign Qa & CRM/Executive & automated & 5 & 5 & csv \\
task\_95 & Board Packet Preparation & CRM/Executive & automated & 5 & 5 & csv \\
task\_96 & Executive Action Item Tracking & CRM/Executive & automated & 5 & 5 & csv \\
task\_97 & Calendar Conflict Resolution & Travel/Office & automated & 5 & 5 & json \\
task\_98 & Travel Itinerary Exception & Travel/Office & automated & 5 & 5 & json \\
task\_99 & Meeting Notes Action Items & Travel/Office & automated & 5 & 5 & json \\
task\_100 & Inbox Rules Classification & Travel/Office & automated & 5 & 5 & json \\
\bottomrule
\end{longtable}
\endgroup

\section{Per-Task Prompt and Output Surfaces}
\label{app:task-output-surfaces}
The following table expands the inventory with the prompt focus and grader-visible output surface for every official task. It is generated from the task markdown files and is intended to make the 100-task composition auditable without requiring reviewers to open each fixture directory.
\begingroup
\scriptsize
\setlength{\tabcolsep}{1.5pt}
\begin{longtable}{L{0.07\textwidth}L{0.20\textwidth}L{0.33\textwidth}L{0.28\textwidth}L{0.08\textwidth}}
\caption{Per-task prompt and output surfaces.}\label{tab:task-output-surfaces}\\
\toprule
\textbf{Task} & \textbf{Name} & \textbf{Prompt focus} & \textbf{Grader-visible output surface} & \textbf{Checks} \\
\midrule
\endfirsthead
\toprule
\textbf{Task} & \textbf{Name} & \textbf{Prompt focus} & \textbf{Grader-visible output surface} & \textbf{Checks} \\
\midrule
\endhead
task\_01 & Batch Data Transform & You have 6 data files in artifact that each contain user records in different formats. & outputs/users\_01.json, outputs/users\_02.json, outputs/users\_03.json ... & automated; 6 criteria \\
task\_02 & Log Analysis & You have 5 log files in artifact from different microservices. & outputs/api\_gateway\_report.json, outputs/auth\_service\_report.json, outputs/notification\_service\_report.json ... & hybrid; 6 criteria \\
task\_03 & API Integration Scaffold & You have 5 API specification files in artifact. & outputs/analytics\_client.py, outputs/inventory\_client.py, outputs/notifications\_client.py ... & automated; 6 criteria \\
task\_04 & Test Generation & You have 6 Python utility modules in artifact. & outputs/test\_data\_utils.py, outputs/test\_date\_utils.py, outputs/test\_file\_utils.py ... & automated; 6 criteria \\
task\_05 & Config Migration v1 to v2 & You have 5 application configuration files in artifact that use a legacy flat key format (v1). & outputs/gateway\_v2.json, outputs/mailer\_v2.json, outputs/scheduler\_v2.json ... & automated; 7 criteria \\
task\_06 & Security Code Review & You have 5 source code files in artifact that contain intentional security vulnerabilities. & outputs/admin\_panel\_review.json, outputs/api\_handler\_review.json, outputs/file\_manager\_review.json ... & hybrid; 7 criteria \\
task\_07 & Document Data Extraction & You have 5 text documents in artifact representing different document types. & outputs/contract\_001.json, outputs/expense\_report\_001.json, outputs/invoice\_XX.json ... & hybrid; 8 criteria \\
task\_08 & Database Schema Operations & You have 5 SQL schema files in artifact representing existing PostgreSQL tables. & outputs/migrate\_orders.sql, outputs/migrate\_products.sql, outputs/migrate\_reviews.sql ... & automated; 7 criteria \\
task\_09 & Excel Analytics Report Generation & You are given 5 regional sales CSV files in artifact. & outputs/report\_central.xlsx, outputs/report\_east.xlsx, outputs/report\_north.xlsx ... & automated; 7 criteria \\
task\_10 & Git Repository History Analysis & You are given 5 bash setup scripts in artifact. & task-specific grader-visible files under outputs/ & automated; 6 criteria \\
task\_11 & Web Page Structured Data Extraction & You are given 5 HTML files in artifact that simulate web pages. & task-specific grader-visible files under outputs/ & hybrid; 7 criteria \\
task\_12 & Word Document Generation & You are given 5 JSON data files in artifact. & task-specific grader-visible files under outputs/ & automated; 6 criteria \\
task\_13 & Multi-Dataset Statistical Analysis Pipeline & You are given 5 groups of related CSV files in artifact. & task-specific grader-visible files under outputs/ & hybrid; 6 criteria \\
task\_14 & Enterprise Email Thread Analysis & Complete enterprise email thread analysis and write grader-visible outputs. & outputs/thread\_XX\_report.json & hybrid; 7 criteria \\
task\_15 & Shell Automation Script Generation & You have 5 automation task specification files in artifact. & outputs/task\_XX\_db\_dump.sh, outputs/task\_XX\_dir\_monitor.sh, outputs/task\_XX\_file\_backup.sh ... & hybrid; 9 criteria \\
task\_16 & Dockerfile and CI Pipeline Generation & You have 5 application specification files in artifact. & outputs/app\_XX/Dockerfile & hybrid; 8 criteria \\
task\_17 & Enterprise Invoice Expense Processing & Complete enterprise invoice expense processing and write grader-visible outputs. & outputs/invoice\_XX\_parsed.json & hybrid; 8 criteria \\
task\_18 & Dependency Security Audit & You have 5 dependency manifest files in artifact from different application stacks. & outputs/gemfile\_audit.json, outputs/go\_audit.json, outputs/package\_audit.json ... & hybrid; 8 criteria \\
task\_19 & Meeting Notes Structured Extraction & Complete meeting notes structured extraction and write grader-visible outputs. & outputs/meeting\_XX\_minutes.json & hybrid; 8 criteria \\
task\_20 & Multi-Environment Config Generation & You have 5 application specification files in artifact. & outputs/data-exporter/config.dev.yml, outputs/ml-inference/config.dev.yml, outputs/notification-service/config.dev.yml ... & automated; 8 criteria \\
task\_21 & System Metrics Anomaly Detection & You have 5 system metrics CSV files in artifact. & outputs/db\_query\_perf\_report.json, outputs/disk\_io\_report.json, outputs/network\_traffic\_report.json ... & hybrid; 8 criteria \\
task\_22 & Finance Ledger Reconciliation & Select the authoritative evidence packet, discard decoys, derive strict JSON case reports. & outputs/case\_XX\_report.json & automated; 8 criteria \\
task\_23 & Subscription Revenue Audit & Select the authoritative evidence packet, discard decoys, derive strict JSON case reports. & outputs/case\_XX\_report.json & automated; 8 criteria \\
task\_24 & Bank KYC Packet Review & Select the authoritative evidence packet, discard decoys, derive strict JSON case reports. & outputs/case\_XX\_report.json & automated; 8 criteria \\
task\_25 & Loan Document Checklist & Select the authoritative evidence packet, discard decoys, derive strict JSON case reports. & outputs/case\_XX\_report.json & automated; 8 criteria \\
task\_26 & Tax Form Consistency & Select the authoritative evidence packet, discard decoys, derive strict JSON case reports. & outputs/case\_XX\_report.json & automated; 8 criteria \\
task\_27 & Contract Clause Extraction & Select the authoritative evidence packet, discard decoys, derive strict JSON case reports. & outputs/case\_XX\_report.json & automated; 8 criteria \\
task\_28 & Discovery Document Tagging & Select the authoritative evidence packet, discard decoys, derive strict JSON case reports. & outputs/case\_XX\_report.json & automated; 8 criteria \\
task\_29 & Policy Exception Mapping & Select the authoritative evidence packet, discard decoys, derive strict JSON case reports. & outputs/case\_XX\_report.json & automated; 8 criteria \\
task\_30 & Nda Obligation Review & Select the authoritative evidence packet, discard decoys, derive strict JSON case reports. & outputs/case\_XX\_report.json & automated; 8 criteria \\
task\_31 & Vendor Contract Risk & Select the authoritative evidence packet, discard decoys, derive strict JSON case reports. & outputs/case\_XX\_report.json & automated; 8 criteria \\
task\_32 & Appointment Referral Triage & Select the authoritative evidence packet, discard decoys, derive strict JSON case reports. & outputs/case\_XX\_report.json & automated; 8 criteria \\
task\_33 & Pharmacy Inventory Reorder & Select the authoritative evidence packet, discard decoys, derive strict JSON case reports. & outputs/case\_XX\_report.json & automated; 8 criteria \\
task\_34 & Clinical Trial Eligibility & Select the authoritative evidence packet, discard decoys, derive strict JSON case reports. & outputs/case\_XX\_report.json & automated; 8 criteria \\
task\_35 & Insurance Prior Auth Review & Select the authoritative evidence packet, discard decoys, derive strict JSON case reports. & outputs/case\_XX\_report.json & automated; 8 criteria \\
task\_36 & Lab Result Followup Queue & Select the authoritative evidence packet, discard decoys, derive strict JSON case reports. & outputs/case\_XX\_report.json & automated; 8 criteria \\
task\_37 & Procurement Bid Scoring & Select the authoritative evidence packet, discard decoys, derive strict JSON case reports. & outputs/case\_XX\_report.json & automated; 8 criteria \\
task\_38 & Purchase Order Three Way Match & Select the authoritative evidence packet, discard decoys, derive strict JSON case reports. & outputs/case\_XX\_report.json & automated; 8 criteria \\
task\_39 & Shipping Exception Resolution & Select the authoritative evidence packet, discard decoys, derive strict JSON case reports. & outputs/case\_XX\_report.json & automated; 8 criteria \\
task\_40 & Warehouse Picklist Validation & Select the authoritative evidence packet, discard decoys, derive strict JSON case reports. & outputs/case\_XX\_report.json & automated; 8 criteria \\
task\_41 & Supplier Risk Packet & Select the authoritative evidence packet, discard decoys, derive strict JSON case reports. & outputs/case\_XX\_report.json & automated; 8 criteria \\
task\_42 & Employee Onboarding Checklist & Select the authoritative evidence packet, discard decoys, derive strict JSON case reports. & outputs/case\_XX\_report.json & automated; 8 criteria \\
task\_43 & Interview Feedback Calibration & Select the authoritative evidence packet, discard decoys, derive strict JSON case reports. & outputs/case\_XX\_report.json & automated; 8 criteria \\
task\_44 & Admissions Packet Screening & Select the authoritative evidence packet, discard decoys, derive strict JSON case reports. & outputs/case\_XX\_report.json & automated; 8 criteria \\
task\_45 & Assignment Rubric Grading & Select the authoritative evidence packet, discard decoys, derive strict JSON case reports. & outputs/case\_XX\_report.json & automated; 8 criteria \\
task\_46 & Training Completion Audit & Select the authoritative evidence packet, discard decoys, derive strict JSON case reports. & outputs/case\_XX\_report.json & automated; 8 criteria \\
task\_47 & Support Ticket Escalation & Select the authoritative evidence packet, discard decoys, derive strict JSON case reports. & outputs/case\_XX\_report.json & automated; 8 criteria \\
task\_48 & Kb Gap Analysis & Select the authoritative evidence packet, discard decoys, derive strict JSON case reports. & outputs/case\_XX\_report.json & automated; 8 criteria \\
task\_49 & App Review Theme Mining & Select the authoritative evidence packet, discard decoys, derive strict JSON case reports. & outputs/case\_XX\_report.json & automated; 8 criteria \\
task\_50 & Bug Report Deduplication & Select the authoritative evidence packet, discard decoys, derive strict JSON case reports. & outputs/case\_XX\_report.json & automated; 8 criteria \\
task\_51 & Call Center Quality Review & Select the authoritative evidence packet, discard decoys, derive strict JSON case reports. & outputs/case\_XX\_report.json & automated; 8 criteria \\
task\_52 & Kubernetes Policy Review & Select the authoritative evidence packet, discard decoys, derive strict JSON case reports. & outputs/case\_XX\_report.json & automated; 8 criteria \\
task\_53 & Terraform Plan Drift & Select the authoritative evidence packet, discard decoys, derive strict JSON case reports. & outputs/case\_XX\_report.json & automated; 8 criteria \\
task\_54 & SLO Burn Rate Analysis & Select the authoritative evidence packet, discard decoys, derive strict JSON case reports. & outputs/case\_XX\_report.json & automated; 8 criteria \\
task\_55 & CI Pipeline Hardening & Select the authoritative evidence packet, discard decoys, derive strict JSON case reports. & outputs/case\_XX\_report.json & automated; 8 criteria \\
task\_56 & Backup Restore Drill & Select the authoritative evidence packet, discard decoys, derive strict JSON case reports. & outputs/case\_XX\_report.json & automated; 8 criteria \\
task\_57 & Security Alert Correlation & Select the authoritative evidence packet, discard decoys, derive strict JSON case reports. & outputs/case\_XX\_report.json & automated; 8 criteria \\
task\_58 & Phishing Report Triage & Select the authoritative evidence packet, discard decoys, derive strict JSON case reports. & outputs/case\_XX\_report.json & automated; 8 criteria \\
task\_59 & Vulnerability Exception Review & Select the authoritative evidence packet, discard decoys, derive strict JSON case reports. & outputs/case\_XX\_report.json & automated; 8 criteria \\
task\_60 & PII Redaction Release & Select the authoritative evidence packet, discard decoys, derive strict JSON case reports. & outputs/case\_XX\_report.json & automated; 8 criteria \\
task\_61 & DSR Request Routing & Select the authoritative evidence packet, discard decoys, derive strict JSON case reports. & outputs/case\_XX\_report.json & automated; 8 criteria \\
task\_62 & SQL Schema Migration Review & Select the authoritative evidence packet, discard decoys, derive strict JSON case reports. & outputs/case\_XX\_report.json & automated; 8 criteria \\
task\_63 & Messy Csv Normalization & Select the authoritative evidence packet, discard decoys, derive strict JSON case reports. & outputs/case\_XX\_report.json & automated; 8 criteria \\
task\_64 & Dashboard Metric Reconciliation & Select the authoritative evidence packet, discard decoys, derive strict JSON case reports. & outputs/case\_XX\_report.json & automated; 8 criteria \\
task\_65 & Data Quality Rule Authoring & Select the authoritative evidence packet, discard decoys, derive strict JSON case reports. & outputs/case\_XX\_report.json & automated; 8 criteria \\
task\_66 & Experiment Results Analysis & Select the authoritative evidence packet, discard decoys, derive strict JSON case reports. & outputs/case\_XX\_report.json & automated; 8 criteria \\
task\_67 & Research Claim Evidence & Select the authoritative evidence packet, discard decoys, derive strict JSON case reports. & outputs/case\_XX\_report.json & automated; 8 criteria \\
task\_68 & Newsroom Fact Check & Select the authoritative evidence packet, discard decoys, derive strict JSON case reports. & outputs/case\_XX\_report.json & automated; 8 criteria \\
task\_69 & Literature Table Extraction & Select the authoritative evidence packet, discard decoys, derive strict JSON case reports. & outputs/case\_XX\_report.json & automated; 8 criteria \\
task\_70 & Public Web Directory Extraction & Select the authoritative evidence packet, discard decoys, derive strict JSON case reports. & outputs/case\_XX\_report.json & automated; 8 criteria \\
task\_71 & Citation Deduplication & Select the authoritative evidence packet, discard decoys, derive strict JSON case reports. & outputs/case\_XX\_report.json & automated; 8 criteria \\
task\_72 & Localization Placeholder Qa & Select the authoritative evidence packet, discard decoys, derive strict JSON case reports. & outputs/case\_XX\_report.json & automated; 8 criteria \\
task\_73 & Glossary Compliance Review & Select the authoritative evidence packet, discard decoys, derive strict JSON case reports. & outputs/case\_XX\_report.json & automated; 8 criteria \\
task\_74 & Release Notes Curation & Select the authoritative evidence packet, discard decoys, derive strict JSON case reports. & outputs/case\_XX\_report.json & automated; 8 criteria \\
task\_75 & Changelog Impact Matrix & Select the authoritative evidence packet, discard decoys, derive strict JSON case reports. & outputs/case\_XX\_report.json & automated; 8 criteria \\
task\_76 & Feature Flag Cleanup Plan & Select the authoritative evidence packet, discard decoys, derive strict JSON case reports. & outputs/case\_XX\_report.json & automated; 8 criteria \\
task\_77 & Ecommerce Catalog Normalization & Select the authoritative evidence packet, discard decoys, derive strict JSON case reports. & outputs/case\_XX\_report.json & automated; 8 criteria \\
task\_78 & Retail Returns Root Cause & Select the authoritative evidence packet, discard decoys, derive strict JSON case reports. & outputs/case\_XX\_report.json & automated; 8 criteria \\
task\_79 & Restaurant Inspection Summary & Select the authoritative evidence packet, discard decoys, derive strict JSON case reports. & outputs/case\_XX\_report.json & automated; 8 criteria \\
task\_80 & Food Delivery Refund Review & Select the authoritative evidence packet, discard decoys, derive strict JSON case reports. & outputs/case\_XX\_report.json & automated; 8 criteria \\
task\_81 & Marketplace Listing Policy & Select the authoritative evidence packet, discard decoys, derive strict JSON case reports. & outputs/case\_XX\_report.json & automated; 8 criteria \\
task\_82 & Facilities Maintenance Prioritization & Select the authoritative evidence packet, discard decoys, derive strict JSON case reports. & outputs/case\_XX\_report.json & automated; 8 criteria \\
task\_83 & Fleet Maintenance Schedule & Select the authoritative evidence packet, discard decoys, derive strict JSON case reports. & outputs/case\_XX\_report.json & automated; 8 criteria \\
task\_84 & Iot Sensor Anomaly Review & Select the authoritative evidence packet, discard decoys, derive strict JSON case reports. & outputs/case\_XX\_report.json & automated; 8 criteria \\
task\_85 & Smart Home Support Diagnosis & Select the authoritative evidence packet, discard decoys, derive strict JSON case reports. & outputs/case\_XX\_report.json & automated; 8 criteria \\
task\_86 & Robotics Workcell Event Report & Select the authoritative evidence packet, discard decoys, derive strict JSON case reports. & outputs/case\_XX\_report.json & automated; 8 criteria \\
task\_87 & Civic Service Request Routing & Select the authoritative evidence packet, discard decoys, derive strict JSON case reports. & outputs/case\_XX\_report.json & automated; 8 criteria \\
task\_88 & Public Records Redaction & Select the authoritative evidence packet, discard decoys, derive strict JSON case reports. & outputs/case\_XX\_report.json & automated; 8 criteria \\
task\_89 & Grant Application Completeness & Select the authoritative evidence packet, discard decoys, derive strict JSON case reports. & outputs/case\_XX\_report.json & automated; 8 criteria \\
task\_90 & Audit Evidence Collection & Select the authoritative evidence packet, discard decoys, derive strict JSON case reports. & outputs/case\_XX\_report.json & automated; 8 criteria \\
task\_91 & Risk Register Rollup & Select the authoritative evidence packet, discard decoys, derive strict JSON case reports. & outputs/case\_XX\_report.json & automated; 8 criteria \\
task\_92 & CRM Pipeline Hygiene & Select the authoritative evidence packet, discard decoys, derive strict JSON case reports. & outputs/case\_XX\_report.json & automated; 8 criteria \\
task\_93 & Sales Forecast Variance & Select the authoritative evidence packet, discard decoys, derive strict JSON case reports. & outputs/case\_XX\_report.json & automated; 8 criteria \\
task\_94 & Marketing Campaign Qa & Select the authoritative evidence packet, discard decoys, derive strict JSON case reports. & outputs/case\_XX\_report.json & automated; 8 criteria \\
task\_95 & Board Packet Preparation & Select the authoritative evidence packet, discard decoys, derive strict JSON case reports. & outputs/case\_XX\_report.json & automated; 8 criteria \\
task\_96 & Executive Action Item Tracking & Select the authoritative evidence packet, discard decoys, derive strict JSON case reports. & outputs/case\_XX\_report.json & automated; 8 criteria \\
task\_97 & Calendar Conflict Resolution & Select the authoritative evidence packet, discard decoys, derive strict JSON case reports. & outputs/case\_XX\_report.json & automated; 8 criteria \\
task\_98 & Travel Itinerary Exception & Select the authoritative evidence packet, discard decoys, derive strict JSON case reports. & outputs/case\_XX\_report.json & automated; 8 criteria \\
task\_99 & Meeting Notes Action Items & Select the authoritative evidence packet, discard decoys, derive strict JSON case reports. & outputs/case\_XX\_report.json & automated; 8 criteria \\
task\_100 & Inbox Rules Classification & Select the authoritative evidence packet, discard decoys, derive strict JSON case reports. & outputs/case\_XX\_report.json & automated; 8 criteria \\
\bottomrule
\end{longtable}
\endgroup

\section{Per-Task Grading Surfaces}
\label{app:task-grading-surfaces}
This table records how each official task is checked. It separates the grading type from the concrete surface that the grader inspects, which is useful because hybrid tasks still require file artifacts and automated tasks may include rich schema or content checks.
\begingroup
\footnotesize
\setlength{\tabcolsep}{3pt}
\begin{longtable}{L{0.07\textwidth}L{0.21\textwidth}L{0.11\textwidth}L{0.16\textwidth}L{0.38\textwidth}}
\caption{Per-task grading surfaces.}\label{tab:task-grading-surfaces}\\
\toprule
\textbf{Task} & \textbf{Name} & \textbf{Grade} & \textbf{Check surface} & \textbf{Representative criteria} \\
\midrule
\endfirsthead
\toprule
\textbf{Task} & \textbf{Name} & \textbf{Grade} & \textbf{Check surface} & \textbf{Representative criteria} \\
\midrule
\endhead
task\_01 & Batch Data Transform & automated & automated code & Each of the 6 output files exists in artifact; Each output file contains valid JSON; Each record conforms to the target schema (all 6 fields present) \\
task\_02 & Log Analysis & hybrid & automated code, LLM rubric, weighted blend & Each of the 5 output report files exists; Each report contains valid JSON; Each report has all required top-level fields \\
task\_03 & API Integration Scaffold & automated & automated code & Each of the 5 output .py files exists; Each file contains valid Python (parseable by ast.parse); Each file contains credential-handling code \\
task\_04 & Test Generation & automated & automated code & Each of the 6 test files exists in artifact; Each file contains valid Python syntax; Each file imports pytest \\
task\_05 & Config Migration v1 to v2 & automated & automated code & All 5 output files exist in outputs/; All output files are valid JSON; All output files use nested structure (no flat dot-notation keys) \\
task\_06 & Security Code Review & hybrid & automated code, LLM rubric, weighted blend & All 5 output files exist in outputs/; All output files are valid JSON; Each report has a "vulnerabilities" array with at least 3 entries \\
task\_07 & Document Data Extraction & hybrid & automated code, LLM rubric, weighted blend & All 5 output files exist in outputs/; All output files are valid JSON; Invoice has line\_items array and grand\_total \\
task\_08 & Database Schema Operations & automated & automated code & All 5 output files exist in outputs/; All files contain valid SQL statements; Each migration contains the required ALTER TABLE statements \\
task\_09 & Excel Analytics Report Generation & automated & automated code & Output .xlsx files exist for all 5 regions; Each workbook has exactly 3 sheets named correctly; Raw Data sheet contains the expected number of rows (20 data rows + header) \\
task\_10 & Git Repository History Analysis & automated & automated code & Output JSON files exist for all 5 repositories; artifact matches expected value exactly; All expected contributors appear in the contributors list \\
task\_11 & Web Page Structured Data Extraction & hybrid & automated code, LLM rubric, weighted blend & All 5 output JSON files exist; Each file contains a list with at least 5 items; Product catalog: prices are numeric, product IDs match source \\
task\_12 & Word Document Generation & automated & automated code & All 5 .docx files exist in outputs/; Files are valid Word documents (openable by python-docx); Each document contains the correct title as a Heading 1 \\
task\_13 & Multi-Dataset Statistical Analysis Pipeline & hybrid & automated code, LLM rubric, weighted blend & All 5 analysis JSON files exist; artifact / artifact / artifact / artifact match expected counts; Numeric statistics are present and of correct type \\
task\_14 & Enterprise Email Thread Analysis & hybrid & automated code, LLM rubric, weighted blend & All 5 output files exist; each report is valid JSON; each report contains all required top-level fields \\
task\_15 & Shell Automation Script Generation & hybrid & automated code, LLM rubric, weighted blend & All 10 output files exist (5 scripts + 5 READMEs); Each artifact file starts with artifact or artifact; Each artifact file contains artifact or artifact \\
task\_16 & Dockerfile and CI Pipeline Generation & hybrid & automated code, LLM rubric, weighted blend & All 10 output files exist (5 Dockerfiles + 5 workflow YAMLs); Each Dockerfile starts with artifact; Multi-stage Dockerfiles (apps 01--04) contain at least 2 artifact instructions \\
task\_17 & Enterprise Invoice Expense Processing & hybrid & automated code, LLM rubric, weighted blend & All 6 output files exist (5 parsed files plus 1 summary); each parsed file is valid JSON; parsed files contain all required top-level fields \\
task\_18 & Dependency Security Audit & hybrid & automated code, LLM rubric, weighted blend & All 5 output files exist; Each report contains valid JSON; Each report has all required top-level fields \\
task\_19 & Meeting Notes Structured Extraction & hybrid & automated code, LLM rubric, weighted blend & All 5 output files exist; each file is valid JSON; each file contains all required top-level fields \\
task\_20 & Multi-Environment Config Generation & automated & automated code & All 15 output YAML files exist; Each YAML file is parseable (valid YAML); artifact has artifact \\
task\_21 & System Metrics Anomaly Detection & hybrid & automated code, LLM rubric, weighted blend & All 5 output report files exist; Each report contains valid JSON; Each report has all required top-level fields \\
task\_22 & Finance Ledger Reconciliation & automated & automated code & All five artifact files exist.; Each report is valid JSON and contains every required field.; The report ignores draft, superseded, invalid-checksum, and decoy packet records. \\
task\_23 & Subscription Revenue Audit & automated & automated code & All five artifact files exist.; Each report is valid JSON and contains every required field.; The report ignores draft, superseded, invalid-checksum, and decoy packet records. \\
task\_24 & Bank KYC Packet Review & automated & automated code & All five artifact files exist.; Each report is valid JSON and contains every required field.; The report ignores draft, superseded, invalid-checksum, and decoy packet records. \\
task\_25 & Loan Document Checklist & automated & automated code & All five artifact files exist.; Each report is valid JSON and contains every required field.; The report ignores draft, superseded, invalid-checksum, and decoy packet records. \\
task\_26 & Tax Form Consistency & automated & automated code & All five artifact files exist.; Each report is valid JSON and contains every required field.; The report ignores draft, superseded, invalid-checksum, and decoy packet records. \\
task\_27 & Contract Clause Extraction & automated & automated code & All five artifact files exist.; Each report is valid JSON and contains every required field.; The report ignores draft, superseded, invalid-checksum, and decoy packet records. \\
task\_28 & Discovery Document Tagging & automated & automated code & All five artifact files exist.; Each report is valid JSON and contains every required field.; The report ignores draft, superseded, invalid-checksum, and decoy packet records. \\
task\_29 & Policy Exception Mapping & automated & automated code & All five artifact files exist.; Each report is valid JSON and contains every required field.; The report ignores draft, superseded, invalid-checksum, and decoy packet records. \\
task\_30 & Nda Obligation Review & automated & automated code & All five artifact files exist.; Each report is valid JSON and contains every required field.; The report ignores draft, superseded, invalid-checksum, and decoy packet records. \\
task\_31 & Vendor Contract Risk & automated & automated code & All five artifact files exist.; Each report is valid JSON and contains every required field.; The report ignores draft, superseded, invalid-checksum, and decoy packet records. \\
task\_32 & Appointment Referral Triage & automated & automated code & All five artifact files exist.; Each report is valid JSON and contains every required field.; The report ignores draft, superseded, invalid-checksum, and decoy packet records. \\
task\_33 & Pharmacy Inventory Reorder & automated & automated code & All five artifact files exist.; Each report is valid JSON and contains every required field.; The report ignores draft, superseded, invalid-checksum, and decoy packet records. \\
task\_34 & Clinical Trial Eligibility & automated & automated code & All five artifact files exist.; Each report is valid JSON and contains every required field.; The report ignores draft, superseded, invalid-checksum, and decoy packet records. \\
task\_35 & Insurance Prior Auth Review & automated & automated code & All five artifact files exist.; Each report is valid JSON and contains every required field.; The report ignores draft, superseded, invalid-checksum, and decoy packet records. \\
task\_36 & Lab Result Followup Queue & automated & automated code & All five artifact files exist.; Each report is valid JSON and contains every required field.; The report ignores draft, superseded, invalid-checksum, and decoy packet records. \\
task\_37 & Procurement Bid Scoring & automated & automated code & All five artifact files exist.; Each report is valid JSON and contains every required field.; The report ignores draft, superseded, invalid-checksum, and decoy packet records. \\
task\_38 & Purchase Order Three Way Match & automated & automated code & All five artifact files exist.; Each report is valid JSON and contains every required field.; The report ignores draft, superseded, invalid-checksum, and decoy packet records. \\
task\_39 & Shipping Exception Resolution & automated & automated code & All five artifact files exist.; Each report is valid JSON and contains every required field.; The report ignores draft, superseded, invalid-checksum, and decoy packet records. \\
task\_40 & Warehouse Picklist Validation & automated & automated code & All five artifact files exist.; Each report is valid JSON and contains every required field.; The report ignores draft, superseded, invalid-checksum, and decoy packet records. \\
task\_41 & Supplier Risk Packet & automated & automated code & All five artifact files exist.; Each report is valid JSON and contains every required field.; The report ignores draft, superseded, invalid-checksum, and decoy packet records. \\
task\_42 & Employee Onboarding Checklist & automated & automated code & All five artifact files exist.; Each report is valid JSON and contains every required field.; The report ignores draft, superseded, invalid-checksum, and decoy packet records. \\
task\_43 & Interview Feedback Calibration & automated & automated code & All five artifact files exist.; Each report is valid JSON and contains every required field.; The report ignores draft, superseded, invalid-checksum, and decoy packet records. \\
task\_44 & Admissions Packet Screening & automated & automated code & All five artifact files exist.; Each report is valid JSON and contains every required field.; The report ignores draft, superseded, invalid-checksum, and decoy packet records. \\
task\_45 & Assignment Rubric Grading & automated & automated code & All five artifact files exist.; Each report is valid JSON and contains every required field.; The report ignores draft, superseded, invalid-checksum, and decoy packet records. \\
task\_46 & Training Completion Audit & automated & automated code & All five artifact files exist.; Each report is valid JSON and contains every required field.; The report ignores draft, superseded, invalid-checksum, and decoy packet records. \\
task\_47 & Support Ticket Escalation & automated & automated code & All five artifact files exist.; Each report is valid JSON and contains every required field.; The report ignores draft, superseded, invalid-checksum, and decoy packet records. \\
task\_48 & Kb Gap Analysis & automated & automated code & All five artifact files exist.; Each report is valid JSON and contains every required field.; The report ignores draft, superseded, invalid-checksum, and decoy packet records. \\
task\_49 & App Review Theme Mining & automated & automated code & All five artifact files exist.; Each report is valid JSON and contains every required field.; The report ignores draft, superseded, invalid-checksum, and decoy packet records. \\
task\_50 & Bug Report Deduplication & automated & automated code & All five artifact files exist.; Each report is valid JSON and contains every required field.; The report ignores draft, superseded, invalid-checksum, and decoy packet records. \\
task\_51 & Call Center Quality Review & automated & automated code & All five artifact files exist.; Each report is valid JSON and contains every required field.; The report ignores draft, superseded, invalid-checksum, and decoy packet records. \\
task\_52 & Kubernetes Policy Review & automated & automated code & All five artifact files exist.; Each report is valid JSON and contains every required field.; The report ignores draft, superseded, invalid-checksum, and decoy packet records. \\
task\_53 & Terraform Plan Drift & automated & automated code & All five artifact files exist.; Each report is valid JSON and contains every required field.; The report ignores draft, superseded, invalid-checksum, and decoy packet records. \\
task\_54 & SLO Burn Rate Analysis & automated & automated code & All five artifact files exist.; Each report is valid JSON and contains every required field.; The report ignores draft, superseded, invalid-checksum, and decoy packet records. \\
task\_55 & CI Pipeline Hardening & automated & automated code & All five artifact files exist.; Each report is valid JSON and contains every required field.; The report ignores draft, superseded, invalid-checksum, and decoy packet records. \\
task\_56 & Backup Restore Drill & automated & automated code & All five artifact files exist.; Each report is valid JSON and contains every required field.; The report ignores draft, superseded, invalid-checksum, and decoy packet records. \\
task\_57 & Security Alert Correlation & automated & automated code & All five artifact files exist.; Each report is valid JSON and contains every required field.; The report ignores draft, superseded, invalid-checksum, and decoy packet records. \\
task\_58 & Phishing Report Triage & automated & automated code & All five artifact files exist.; Each report is valid JSON and contains every required field.; The report ignores draft, superseded, invalid-checksum, and decoy packet records. \\
task\_59 & Vulnerability Exception Review & automated & automated code & All five artifact files exist.; Each report is valid JSON and contains every required field.; The report ignores draft, superseded, invalid-checksum, and decoy packet records. \\
task\_60 & PII Redaction Release & automated & automated code & All five artifact files exist.; Each report is valid JSON and contains every required field.; The report ignores draft, superseded, invalid-checksum, and decoy packet records. \\
task\_61 & DSR Request Routing & automated & automated code & All five artifact files exist.; Each report is valid JSON and contains every required field.; The report ignores draft, superseded, invalid-checksum, and decoy packet records. \\
task\_62 & SQL Schema Migration Review & automated & automated code & All five artifact files exist.; Each report is valid JSON and contains every required field.; The report ignores draft, superseded, invalid-checksum, and decoy packet records. \\
task\_63 & Messy Csv Normalization & automated & automated code & All five artifact files exist.; Each report is valid JSON and contains every required field.; The report ignores draft, superseded, invalid-checksum, and decoy packet records. \\
task\_64 & Dashboard Metric Reconciliation & automated & automated code & All five artifact files exist.; Each report is valid JSON and contains every required field.; The report ignores draft, superseded, invalid-checksum, and decoy packet records. \\
task\_65 & Data Quality Rule Authoring & automated & automated code & All five artifact files exist.; Each report is valid JSON and contains every required field.; The report ignores draft, superseded, invalid-checksum, and decoy packet records. \\
task\_66 & Experiment Results Analysis & automated & automated code & All five artifact files exist.; Each report is valid JSON and contains every required field.; The report ignores draft, superseded, invalid-checksum, and decoy packet records. \\
task\_67 & Research Claim Evidence & automated & automated code & All five artifact files exist.; Each report is valid JSON and contains every required field.; The report ignores draft, superseded, invalid-checksum, and decoy packet records. \\
task\_68 & Newsroom Fact Check & automated & automated code & All five artifact files exist.; Each report is valid JSON and contains every required field.; The report ignores draft, superseded, invalid-checksum, and decoy packet records. \\
task\_69 & Literature Table Extraction & automated & automated code & All five artifact files exist.; Each report is valid JSON and contains every required field.; The report ignores draft, superseded, invalid-checksum, and decoy packet records. \\
task\_70 & Public Web Directory Extraction & automated & automated code & All five artifact files exist.; Each report is valid JSON and contains every required field.; The report ignores draft, superseded, invalid-checksum, and decoy packet records. \\
task\_71 & Citation Deduplication & automated & automated code & All five artifact files exist.; Each report is valid JSON and contains every required field.; The report ignores draft, superseded, invalid-checksum, and decoy packet records. \\
task\_72 & Localization Placeholder Qa & automated & automated code & All five artifact files exist.; Each report is valid JSON and contains every required field.; The report ignores draft, superseded, invalid-checksum, and decoy packet records. \\
task\_73 & Glossary Compliance Review & automated & automated code & All five artifact files exist.; Each report is valid JSON and contains every required field.; The report ignores draft, superseded, invalid-checksum, and decoy packet records. \\
task\_74 & Release Notes Curation & automated & automated code & All five artifact files exist.; Each report is valid JSON and contains every required field.; The report ignores draft, superseded, invalid-checksum, and decoy packet records. \\
task\_75 & Changelog Impact Matrix & automated & automated code & All five artifact files exist.; Each report is valid JSON and contains every required field.; The report ignores draft, superseded, invalid-checksum, and decoy packet records. \\
task\_76 & Feature Flag Cleanup Plan & automated & automated code & All five artifact files exist.; Each report is valid JSON and contains every required field.; The report ignores draft, superseded, invalid-checksum, and decoy packet records. \\
task\_77 & Ecommerce Catalog Normalization & automated & automated code & All five artifact files exist.; Each report is valid JSON and contains every required field.; The report ignores draft, superseded, invalid-checksum, and decoy packet records. \\
task\_78 & Retail Returns Root Cause & automated & automated code & All five artifact files exist.; Each report is valid JSON and contains every required field.; The report ignores draft, superseded, invalid-checksum, and decoy packet records. \\
task\_79 & Restaurant Inspection Summary & automated & automated code & All five artifact files exist.; Each report is valid JSON and contains every required field.; The report ignores draft, superseded, invalid-checksum, and decoy packet records. \\
task\_80 & Food Delivery Refund Review & automated & automated code & All five artifact files exist.; Each report is valid JSON and contains every required field.; The report ignores draft, superseded, invalid-checksum, and decoy packet records. \\
task\_81 & Marketplace Listing Policy & automated & automated code & All five artifact files exist.; Each report is valid JSON and contains every required field.; The report ignores draft, superseded, invalid-checksum, and decoy packet records. \\
task\_82 & Facilities Maintenance Prioritization & automated & automated code & All five artifact files exist.; Each report is valid JSON and contains every required field.; The report ignores draft, superseded, invalid-checksum, and decoy packet records. \\
task\_83 & Fleet Maintenance Schedule & automated & automated code & All five artifact files exist.; Each report is valid JSON and contains every required field.; The report ignores draft, superseded, invalid-checksum, and decoy packet records. \\
task\_84 & Iot Sensor Anomaly Review & automated & automated code & All five artifact files exist.; Each report is valid JSON and contains every required field.; The report ignores draft, superseded, invalid-checksum, and decoy packet records. \\
task\_85 & Smart Home Support Diagnosis & automated & automated code & All five artifact files exist.; Each report is valid JSON and contains every required field.; The report ignores draft, superseded, invalid-checksum, and decoy packet records. \\
task\_86 & Robotics Workcell Event Report & automated & automated code & All five artifact files exist.; Each report is valid JSON and contains every required field.; The report ignores draft, superseded, invalid-checksum, and decoy packet records. \\
task\_87 & Civic Service Request Routing & automated & automated code & All five artifact files exist.; Each report is valid JSON and contains every required field.; The report ignores draft, superseded, invalid-checksum, and decoy packet records. \\
task\_88 & Public Records Redaction & automated & automated code & All five artifact files exist.; Each report is valid JSON and contains every required field.; The report ignores draft, superseded, invalid-checksum, and decoy packet records. \\
task\_89 & Grant Application Completeness & automated & automated code & All five artifact files exist.; Each report is valid JSON and contains every required field.; The report ignores draft, superseded, invalid-checksum, and decoy packet records. \\
task\_90 & Audit Evidence Collection & automated & automated code & All five artifact files exist.; Each report is valid JSON and contains every required field.; The report ignores draft, superseded, invalid-checksum, and decoy packet records. \\
task\_91 & Risk Register Rollup & automated & automated code & All five artifact files exist.; Each report is valid JSON and contains every required field.; The report ignores draft, superseded, invalid-checksum, and decoy packet records. \\
task\_92 & CRM Pipeline Hygiene & automated & automated code & All five artifact files exist.; Each report is valid JSON and contains every required field.; The report ignores draft, superseded, invalid-checksum, and decoy packet records. \\
task\_93 & Sales Forecast Variance & automated & automated code & All five artifact files exist.; Each report is valid JSON and contains every required field.; The report ignores draft, superseded, invalid-checksum, and decoy packet records. \\
task\_94 & Marketing Campaign Qa & automated & automated code & All five artifact files exist.; Each report is valid JSON and contains every required field.; The report ignores draft, superseded, invalid-checksum, and decoy packet records. \\
task\_95 & Board Packet Preparation & automated & automated code & All five artifact files exist.; Each report is valid JSON and contains every required field.; The report ignores draft, superseded, invalid-checksum, and decoy packet records. \\
task\_96 & Executive Action Item Tracking & automated & automated code & All five artifact files exist.; Each report is valid JSON and contains every required field.; The report ignores draft, superseded, invalid-checksum, and decoy packet records. \\
task\_97 & Calendar Conflict Resolution & automated & automated code & All five artifact files exist.; Each report is valid JSON and contains every required field.; The report ignores draft, superseded, invalid-checksum, and decoy packet records. \\
task\_98 & Travel Itinerary Exception & automated & automated code & All five artifact files exist.; Each report is valid JSON and contains every required field.; The report ignores draft, superseded, invalid-checksum, and decoy packet records. \\
task\_99 & Meeting Notes Action Items & automated & automated code & All five artifact files exist.; Each report is valid JSON and contains every required field.; The report ignores draft, superseded, invalid-checksum, and decoy packet records. \\
task\_100 & Inbox Rules Classification & automated & automated code & All five artifact files exist.; Each report is valid JSON and contains every required field.; The report ignores draft, superseded, invalid-checksum, and decoy packet records. \\
\bottomrule
\end{longtable}
\endgroup

\section{Task Families and Fixture Mechanics}
\label{app:task-families}
The current suite combines the original heterogeneous seed tasks with generated hard-mode families. The generated families share a common evidence-packet protocol, but each family changes the business action, output schema, and grader-visible decision surface. This is intended to make label diversity insufficient: an agent must handle repeated mechanics while still mapping them to distinct artifacts.
\begingroup
\small
\setlength{\tabcolsep}{3pt}
\begin{longtable}{L{0.18\textwidth}L{0.09\textwidth}L{0.13\textwidth}L{0.52\textwidth}}
\caption{Task-family summary.}\label{tab:task-family-summary}\\
\toprule
\textbf{Family} & \textbf{Tasks} & \textbf{Sub-problems} & \textbf{Fixture and output mechanics} \\
\midrule
\endfirsthead
\toprule
\textbf{Family} & \textbf{Tasks} & \textbf{Sub-problems} & \textbf{Fixture and output mechanics} \\
\midrule
\endhead
Original seed tasks & 21 & 107 & Early heterogeneous tasks covering code, logs, documents, office files, shell automation, CI, dependencies, and metrics; these use task-specific fixtures and graders rather than the shared hard-mode packet protocol. Tasks: 01, 02, 03, 04, 05, 06, 07, 08, 09, 10, 11, 12, 13, 14, 15, 16, 17, 18, 19, 20, 21. \\
CRM/Executive & 5 & 25 & Pipeline hygiene, forecast variance, campaign QA, board packets, and executive action tracking with repeated business-report outputs. Tasks: 92, 93, 94, 95, 96. \\
Commerce/Food & 5 & 25 & Delivery refunds, inspections, marketplace policies, maintenance-like retail cases, and supportable operational decisions over generated local fixtures. Tasks: 77, 78, 79, 80, 81. \\
Data/Analytics & 5 & 25 & Normalization, reconciliation, rule-authoring, experiment, and extraction tasks that stress repeatable data derivation under decoys and stale revisions. Tasks: 62, 63, 64, 65, 66. \\
DevOps/SRE & 5 & 25 & Policy, drift, burn-rate, pipeline-hardening, and restore-drill tasks with infrastructure-oriented evidence channels and output artifacts. Tasks: 52, 53, 54, 55, 56. \\
Facilities/IoT & 5 & 25 & Facilities, fleet, sensor, smart-home, and workcell event tasks that derive maintenance or anomaly reports from selected evidence packets. Tasks: 82, 83, 84, 85, 86. \\
Finance & 5 & 25 & Ledger, revenue, loan, tax, and KYC-style reconciliation with selected evidence packets, stale revisions, numeric deltas, list edits, and boolean compliance gates. Tasks: 22, 23, 24, 25, 26. \\
HR/Education & 5 & 25 & Onboarding, interview, admissions, grading, and completion-audit cases requiring structured checklist or rubric outputs from decoy-rich evidence. Tasks: 42, 43, 44, 45, 46. \\
Healthcare & 5 & 25 & Clinical-administration tasks such as triage, pharmacy reorder, trial eligibility, prior authorization, and follow-up queues with strict output schemas. Tasks: 32, 33, 34, 35, 36. \\
Legal & 5 & 25 & Clause, discovery, policy-exception, NDA, and contract-risk review tasks that require choosing authoritative records and writing structured legal-operation outputs. Tasks: 27, 28, 29, 30, 31. \\
Localization/Release & 5 & 25 & Placeholder QA, release notes, flag cleanup, catalog normalization, and returns analysis tasks with structured release or commerce-adjacent artifacts. Tasks: 72, 73, 74, 75, 76. \\
Procurement/Logistics & 5 & 25 & Supplier, bid, purchase-order, warehouse, and shipping cases built around conflicting packets, exception lists, and derived routing or scoring reports. Tasks: 37, 38, 39, 40, 41. \\
Public Sector/Audit & 5 & 25 & Service routing, records redaction, grant completeness, evidence collection, and risk register tasks using synthetic civic/audit fixtures. Tasks: 87, 88, 89, 90, 91. \\
Research/Media & 5 & 25 & Claim checking, table extraction, directory extraction, citation deduplication, and glossary compliance with evidence selection and citation-like outputs. Tasks: 67, 68, 69, 70, 71. \\
Security/Privacy & 5 & 25 & Alert, phishing, exception, release-redaction, and request-routing tasks that test privacy/security triage without using real sensitive data. Tasks: 57, 58, 59, 60, 61. \\
Support/Product & 5 & 25 & Ticket, knowledge-base, app-review, bug-deduplication, and call-quality tasks that convert repeated records into grader-visible JSON reports. Tasks: 47, 48, 49, 50, 51. \\
Travel/Office & 4 & 20 & Calendar, itinerary, meeting-action, and inbox-rule cases that convert office coordination evidence into structured decisions. Tasks: 97, 98, 99, 100. \\
\bottomrule
\end{longtable}
\endgroup

\section{Generated Family Design Matrix}
\label{app:family-design-matrix}

\Cref{tab:family-design-matrix} expands the family summary with the repeated structure that is intended to make skill creation plausible, and the anti-shortcut mechanism that prevents a shallow schema-only answer from satisfying the grader. The matrix is written at the family level rather than the task level because generated tasks within a family share the same evidence-selection contract while changing the concrete business operation and required output fields.
\begingroup
\small
\setlength{\tabcolsep}{3pt}
\begin{longtable}{L{0.18\textwidth}L{0.24\textwidth}L{0.27\textwidth}L{0.23\textwidth}}
\caption{Design matrix for generated hard-mode task families.}\label{tab:family-design-matrix}\\
\toprule
\textbf{Family} & \textbf{Repeated structure} & \textbf{Anti-shortcut mechanism} & \textbf{Grader-visible artifact} \\
\midrule
\endfirsthead
\toprule
\textbf{Family} & \textbf{Repeated structure} & \textbf{Anti-shortcut mechanism} & \textbf{Grader-visible artifact} \\
\midrule
\endhead
CRM/Executive & Repeated business-status packets must be reconciled into executive-ready decisions. & Decoy opportunities, stale campaign revisions, and conflicting forecast evidence make visible totals unreliable. & Structured pipeline, forecast, campaign, board, or action-item JSON reports. \\
Commerce/Food & Repeated commerce cases map operational evidence to refund, inspection, marketplace, or returns decisions. & Draft and superseded packets expose plausible but wrong SKUs, refund IDs, or policy codes. & Per-case operational JSON reports under \path{outputs/}. \\
Data/Analytics & Repeated analytics records require applying numeric, list, and text derivation channels consistently. & Schema-only outputs fail because selected packets must be verified before computing rows, rules, or experiment winners. & Strict JSON reports with derived counts, deltas, failures, and winners. \\
DevOps/SRE & Repeated infrastructure cases ask for drift, burn-rate, policy, pipeline, or restore conclusions. & Stale revisions and decoy resource identifiers make direct copying from fixtures unsafe. & JSON reports that encode affected resources, exception IDs, and remediation signals. \\
Facilities/IoT & Repeated sensor, fleet, smart-home, and workcell packets require deriving maintenance or anomaly actions. & Conflicting packet states and boolean gates separate approved evidence from noisy telemetry. & Prioritization, diagnosis, schedule, or event-summary JSON reports. \\
Finance & Repeated finance cases share ledger-style aggregation, exception lists, tax-like numeric fields, and compliance gates. & Invalid checksums, superseded packets, and decoy ledgers prevent copying visible totals. & Reconciliation or packet-review JSON reports. \\
HR/Education & Repeated people-process cases map evidence to checklists, rubrics, screening, or calibration decisions. & Decoy records include plausible missing items or scores that must be ignored unless final and selected. & Checklist, rubric, admissions, interview, or completion-audit JSON reports. \\
Healthcare & Repeated clinical-administration cases derive triage, reorder, eligibility, authorization, or follow-up decisions. & Synthetic healthcare-like evidence avoids real patient data while still requiring selected-packet derivation. & Strict healthcare-administration JSON reports. \\
Legal & Repeated legal-operation cases extract obligations, clauses, exceptions, or contract risks. & Draft clauses and stale revisions test whether the agent follows packet provenance instead of surface wording. & Clause, discovery, NDA, policy, or risk JSON outputs. \\
Localization/Release & Repeated release-management cases reconcile placeholders, glossary issues, flags, changelog impacts, or catalog data. & Multiple revisions and alias/remove list actions require ordered list derivation rather than keyword matching. & Release, localization, flag, changelog, or catalog-normalization JSON reports. \\
Procurement/Logistics & Repeated sourcing and logistics cases derive vendors, purchase-order exceptions, warehouse issues, or shipment actions. & Competing packet weights and revisions expose believable but wrong vendor and exception values. & Procurement, warehouse, and shipping JSON reports. \\
Public Sector/Audit & Repeated public-service and audit packets require routing, redaction, evidence, grant, or risk decisions. & Synthetic public-sector records contain missing-evidence decoys and privacy-like fields without real PII. & Civic-routing, redaction, grant, evidence, and risk-register JSON reports. \\
Research/Media & Repeated research and media cases derive evidence tables, citations, claims, facts, or glossary compliance. & Text candidates and list edits force deterministic selection rather than free-form summarization. & Claim, fact-check, extraction, citation, or compliance JSON reports. \\
Security/Privacy & Repeated security and privacy cases derive incidents, redactions, phishing, exceptions, or DSR routing. & Decoy indicators and boolean gates separate final selected security evidence from raw noisy records. & Alert, privacy, phishing, exception, or release-redaction JSON reports. \\
Support/Product & Repeated support and product cases derive escalations, KB gaps, app-review themes, bug clusters, or call quality. & Alias and remove actions test whether the agent can update lists rather than aggregate all visible labels. & Support/product JSON reports with routed or clustered decisions. \\
Travel/Office & Repeated office-coordination cases derive conflicts, itinerary exceptions, meeting actions, or inbox classifications. & Stale calendar or inbox-rule revisions make the latest visible item insufficient without packet selection. & Calendar, travel, meeting, or inbox-rule JSON reports. \\
\bottomrule
\end{longtable}
\endgroup

\section{Hard-Mode Evidence Protocol}
\label{app:hard-mode-protocol}
Generated hard-mode tasks are written so that visible values in a fixture are not necessarily authoritative. Each case contains a packet manifest plus evidence records. The agent must select the approved packet with no superseding packet and a valid checksum equal to the first 16 hexadecimal characters of the hash of the benchmark salt, task id, case id, packet id, and nonce. If more than one packet remains, it chooses the highest revision, then the highest source weight, then the lowest packet identifier. Only final records from the selected packet may drive the output.
\begin{quote}\scriptsize\ttfamily
\detokenize{sha256(evoclawbench-difficulty-hardening-20260524-v4|}\par
\detokenize{<task_id>|<case_id>|<packet_id>|<nonce>)}\par
\end{quote}
\begin{center}
\small
\captionof{table}{Evidence-channel derivation rules used by generated hard-mode tasks.}
\label{tab:evidence-channel-rules}
\begin{tabular}{L{0.16\textwidth}L{0.18\textwidth}L{0.56\textwidth}}
\toprule
\textbf{Output type} & \textbf{Record field} & \textbf{Derivation rule} \\
\midrule
numeric & numeric\_delta & Apply signed amount\_minor / scale; subtract when operator is subtract; round non-count values to two decimals. \\
list & list\_action & Apply include, remove, and alias rows in revision order; emit sorted unique strings. \\
dict & dict\_delta & Sum deltas by bucket and omit zero-valued buckets. \\
boolean & boolean\_gate & Emit true only when every selected gate has observed equal to expected. \\
text & text\_candidate & Choose the candidate maximizing score - penalty; break ties lexicographically. \\
\bottomrule
\end{tabular}
\end{center}
The protocol appears across JSON, YAML, CSV, text, and HTML fixtures. JSON and YAML expose manifest and record arrays directly; CSV fixtures use section-tagged rows; text fixtures use JSON lines; HTML fixtures embed JSON in script blocks. The grader checks the derived files under \path{outputs/} and does not accept a final natural-language explanation as a substitute for those artifacts.

\section{Representative Output Schemas}
\label{app:output-schemas}
For generated hard-mode tasks, the prompt names the exact JSON fields that must appear in each case report. The table lists one representative schema per generated family; seed tasks use task-specific artifact formats such as scripts, workbooks, SQL migrations, Dockerfiles, CI files, or extraction JSON.
\begingroup
\small
\setlength{\tabcolsep}{4pt}
\renewcommand{\arraystretch}{1.08}
\begin{longtable}{L{0.20\textwidth}L{0.23\textwidth}L{0.47\textwidth}}
\caption{Representative output schemas for generated hard-mode families.}\label{tab:representative-output-schemas}\\
\toprule
\textbf{Family} & \textbf{Representative task} & \textbf{Required output fields} \\
\midrule
\endfirsthead
\toprule
\textbf{Family} & \textbf{Representative task} & \textbf{Required output fields} \\
\midrule
\endhead
CRM/Executive & task\_92: CRM Pipeline Hygiene & stale\_opportunities, forecast\_delta, campaign\_errors, action\_items, executive\_summary \\
Commerce/Food & task\_77: Ecommerce Catalog Normalization & normalized\_skus, refund\_ids, inspection\_score, policy\_violations, recommended\_action \\
Data/Analytics & task\_62: SQL Schema Migration Review & row\_count, quality\_failures, metric\_delta, rule\_ids, experiment\_winner \\
DevOps/SRE & task\_52: Kubernetes Policy Review & policy\_violations, required\_changes, slo\_status, backup\_passed, risk\_score \\
Facilities/IoT & task\_82: Facilities Maintenance Prioritization & priority\_assets, maintenance\_due, anomaly\_ids, diagnostic\_codes, dispatch\_required \\
Finance & task\_22: Finance Ledger Reconciliation & ledger\_total, exception\_ids, currencies, tax\_total, balanced \\
HR/Education & task\_42: Employee Onboarding Checklist & completion\_rate, missing\_items, calibrated\_scores, assigned\_track, intervention\_ids \\
Healthcare & task\_32: Appointment Referral Triage & eligible\_ids, routing\_queue, followup\_ids, stockout\_ids, urgent\_count \\
Legal & task\_27: Contract Clause Extraction & clause\_labels, missing\_fields, high\_risk\_terms, parties, effective\_date \\
Localization/Release & task\_72: Localization Placeholder Qa & placeholder\_errors, glossary\_violations, release\_sections, flag\_actions, publish\_ready \\
Procurement/Logistics & task\_37: Procurement Bid Scoring & selected\_vendor, match\_exceptions, late\_shipments, risk\_suppliers, savings\_estimate \\
Public Sector/Audit & task\_87: Civic Service Request Routing & routing\_queue, redactions\_required, missing\_evidence, risk\_summary, compliant \\
Research/Media & task\_67: Research Claim Evidence & supported\_claims, unsupported\_claims, source\_count, duplicate\_citations, confidence \\
Security/Privacy & task\_57: Security Alert Correlation & incident\_ids, redacted\_fields, raw\_pii\_present, approved\_exceptions, severity\_counts \\
Support/Product & task\_47: Support Ticket Escalation & priority\_queue, duplicate\_groups, themes, sla\_breaches, reply\_template \\
Travel/Office & task\_97: Calendar Conflict Resolution & conflicts, exceptions, action\_items, rule\_labels, resolved \\
\bottomrule
\end{longtable}
\endgroup

\section{Suite Distributions}
\label{app:suite-distributions}
\begingroup
\begin{center}
\small
\setlength{\tabcolsep}{6pt}
\renewcommand{\arraystretch}{1.08}
\begin{minipage}[t]{0.34\textwidth}
\centering
\captionof{table}{Grading types.}
\label{tab:suite-distributions}
\begin{tabular}{@{}lr@{}}
\toprule
\textbf{Type} & \textbf{Tasks} \\
\midrule
automated & 88 \\
hybrid & 12 \\
\bottomrule
\end{tabular}
\end{minipage}
\hspace{0.08\textwidth}
\begin{minipage}[t]{0.34\textwidth}
\centering
\captionof{table}{Sub-problem counts.}
\label{tab:subproblem-distribution}
\begin{tabular}{@{}lr@{}}
\toprule
\textbf{Sub-problems} & \textbf{Tasks} \\
\midrule
5 & 98 \\
6 & 2 \\
\bottomrule
\end{tabular}
\end{minipage}

\vspace{0.75\baselineskip}

\captionof{table}{Fixture extensions.}
\label{tab:fixture-extension-distribution}
\begin{tabular}{@{}lr@{\hspace{2em}}lr@{\hspace{2em}}lr@{}}
\toprule
\textbf{Ext.} & \textbf{Files} & \textbf{Ext.} & \textbf{Files} & \textbf{Ext.} & \textbf{Files} \\
\midrule
json & 228 & csv & 128 & yaml & 56 \\
txt & 37 & html & 30 & py & 9 \\
log & 5 & sh & 5 & sql & 5 \\
xml & 3 & tsv & 2 & go & 1 \\
js & 1 & mod & 1 & none & 1 \\
\bottomrule
\end{tabular}
\end{center}
\endgroup

\section{Skill Artifact Structure}
\label{app:skill-artifact}
Generated skill directories are discovered under \path{skills/}. The seeded \path{skills/skill-creator} bundle is excluded from created-skill metrics. A generated skill must include \path{SKILL.md}; optional scripts and references are counted when present. Before reuse execution, non-seeded skill files are hashed. The same files are hashed after execution, and any add, delete, or content change is recorded as a mutation violation for that phase.
\begin{center}
\includegraphics[width=0.86\textwidth]{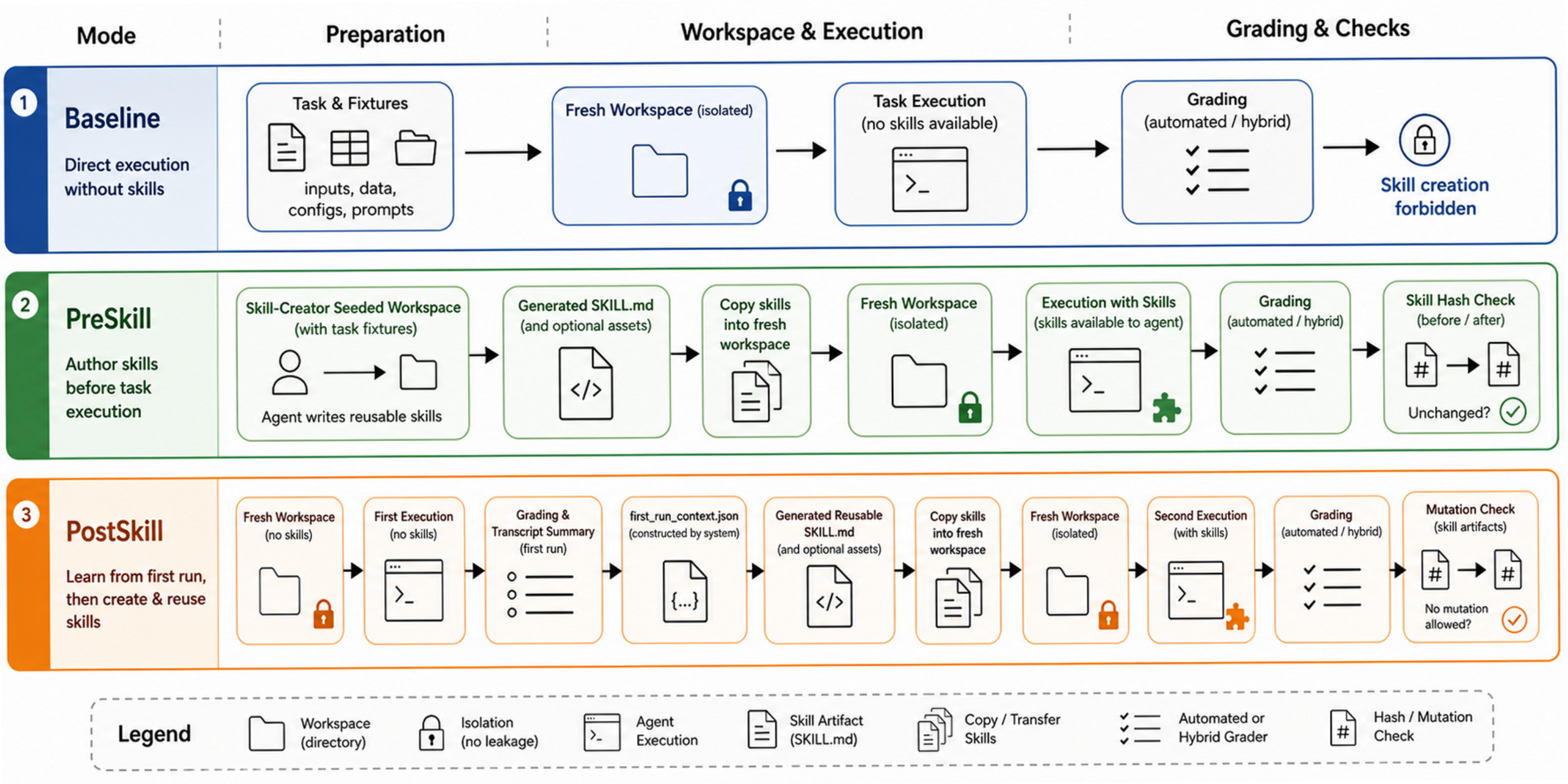}
\captionof{figure}{Three-mode evaluation protocol. \baseline{} directly executes a task without skill creation; \preskill{} first creates task-specific skills and then reuses them in a fresh workspace; \postskill{} summarizes reusable skills from first-run evidence before a second execution. Skill reuse phases are checked for mutation violations.}
\label{fig:protocol}
\end{center}

\section{Result JSON Schema}
\label{app:json}
The aggregate output JSON contains \texttt{baseline\_results}, \texttt{preskill\_results}, \texttt{postskill\_results}, and \texttt{metrics}. The table below summarizes the fields used for benchmark reporting.
\begin{center}
\footnotesize
\captionof{table}{Result JSON fields used for benchmark reporting.}
\label{tab:result-json-fields}
\begin{tabular}{L{0.30\textwidth}L{0.25\textwidth}L{0.35\textwidth}}
\toprule
\textbf{Object} & \textbf{Fields} & \textbf{Use in paper} \\
\midrule
top-level & version, timestamp, benchmark, model, runtime, run\_id, judge model & Identifies the benchmark run, runtime configuration, and grading configuration. \\
top-level & mode, environment, workers, runs\_per\_task & Execution controls for a benchmark run. \\
results & baseline\_results & One graded result object per loaded task for direct execution. \\
results & preskill\_results & One graded result object per loaded task after preskill skill creation and reuse execution. \\
results & postskill\_results & One graded result object per loaded task after first execution, summary, and second execution. \\
metrics.execution\_only & mean\_scores, ratios\_vs\_baseline, usage, efficiency\_vs\_baseline & Final execution phase quality and resource comparisons. \\
metrics.end\_to\_end & mean\_scores, ratios\_vs\_baseline, usage, efficiency\_vs\_baseline & Full workflow quality and overhead comparisons. \\
metrics.created\_skills & preskill\_count, postskill\_count, preskill, postskill & Skill counts and per-skill metadata collected from workspaces. \\
metrics.skill\_quality & preskill, postskill & Heuristic quality summaries for generated skills. \\
metrics.skill\_mutation\_violations & preskill, postskill & Counts reuse executions that added, removed, or edited skill files. \\
\bottomrule
\end{tabular}
\end{center}

\section{Result Validity Checklist}
\label{app:checklist}
Before adding a result row, record the model, runtime, execution mode, environment, worker count, judge model, loaded task count, official task count, mean scores for all modes, created-skill counts, and mutation checks.
Runs should be reported consistently across \baseline{}, \preskill{}, and \postskill{} so that score, cost, and skill-reuse comparisons refer to the same benchmark protocol.

\end{document}